\pdfoutput=1
\documentclass{article}
\usepackage{arxiv}

\usepackage[utf8]{inputenc}
\usepackage[T1]{fontenc}
\usepackage{natbib}
\setcitestyle{authoryear,round,citesep={;},aysep={,},yysep={;}}
\usepackage{url}
\usepackage{booktabs}
\usepackage{amssymb}
\usepackage{nicefrac}
\usepackage{microtype}
\usepackage{xcolor}

\usepackage[all]{nowidow}
\usepackage{algorithm}
\usepackage{algorithmic}
\usepackage{graphicx}
\usepackage{adjustbox}
\usepackage{placeins}
\usepackage{amsmath}
\usepackage{etoolbox}
\usepackage{mdframed}
\usepackage{enumitem}
\usepackage{multirow}
\usepackage{amsthm}
\usepackage{mathtools}
\usepackage{hyperref}

\newmdtheoremenv{example}{Example}[section]
\newmdtheoremenv{observation}{Observation}[section]
\newmdtheoremenv{proposition}{Proposition}[section]

\newcommand{\State}{\STATE}
\newcommand{\For}{\FOR}
\newcommand{\EndFor}{\ENDFOR}

\newcommand{\While}{\WHILE}
\newcommand{\EndWhile}{\ENDWHILE}
\newcommand{\Require}{\REQUIRE}
\newcommand{\Ensure}{\ENSURE}
\newcommand{\Return}{\STATE \textbf{return}}

\title{Neural Tractability via Structure: Learning-Augmented Algorithms for Graph Combinatorial Optimization}

\author{Jialiang Li \qquad Weitong Chen \qquad Mingyu Guo \\
Adelaide University \\
\texttt{\{j.li, weitong.chen, mingyu.guo\}@adelaide.edu.au}
}

\date{}

\renewcommand{\shorttitle}{}

\hypersetup{
pdftitle={Neural Tractability via Structure: Learning-Augmented Algorithms for Graph Combinatorial Optimization},
pdfauthor={Jialiang Li, Weitong Chen, Mingyu Guo},
}

\begin{document}

\maketitle

\begin{abstract}
Neural solvers provide fast solutions to graph combinatorial optimization
problems, but training alone does not guarantee solution quality. Exact
methods guarantee optimality but can be prohibitively expensive.
We propose Neural Fixed-Parameter Tractable (\textsc{N-FPT}), a
\emph{neural-model-agnostic} framework that uses parameterized algorithms
to improve neural solutions without retraining and guide learning.
It restricts neural advice to a treewidth modulator and completes the
bounded-treewidth remainder exactly. It returns and certifies the
\emph{best achievable solution} under any valid advice, improving the neural
solution whenever a better compatible completion exists. We prove this
guarantee and global optimality under perfect advice. Its incremental-confidence
\textsc{N-FPT} (\textsc{NIC-FPT}) and randomized-deferral \textsc{N-FPT}
(\textsc{NRD-FPT}) variants consolidate neural advice.
Beyond inference, \emph{\textsc{N-FPT} feedback} becomes available as neural
decisions make exact completion tractable. From that state onward, it
supplies what final solution rewards lack: the \emph{first decision that
rules out every optimal completion} and alternatives that preserve one.
One exact computation supplies reusable guidance for continuations from
that state, even when neural samples miss its optimum.
Experiments across graph combinatorial optimization problems demonstrate
solution-quality gains with different neural models and robustness to
graph-size and distribution shifts. The learned completion model reaches
optimal completions more often than reinforcement learning (RL) alone,
without an exact solver at inference. \textsc{N-FPT} thus uses graph structure
to improve neural solutions and guide learning toward optimal completion.
\end{abstract}

\section{Introduction} \label{sec:introduction}
Graph combinatorial optimization seeks the best discrete choices under graph
constraints: for example, selecting as many mutually nonadjacent vertices as
possible or partitioning vertices to maximize the number of crossing edges.
Many such problems are NP-hard. Exact methods guarantee optimality but can
be prohibitively expensive, while neural solvers offer fast predictions after
training without comparable guarantees~\citep{zhang2023letflowstellsolving}.
Parameterized algorithms offer a way to use \emph{graph structure} to make exact
computation tractable~\citep{cygan2015parameterized,fomin2019kernelization}.

\begin{figure}[!ht]
  \centering
  \includegraphics[width=0.423\linewidth]{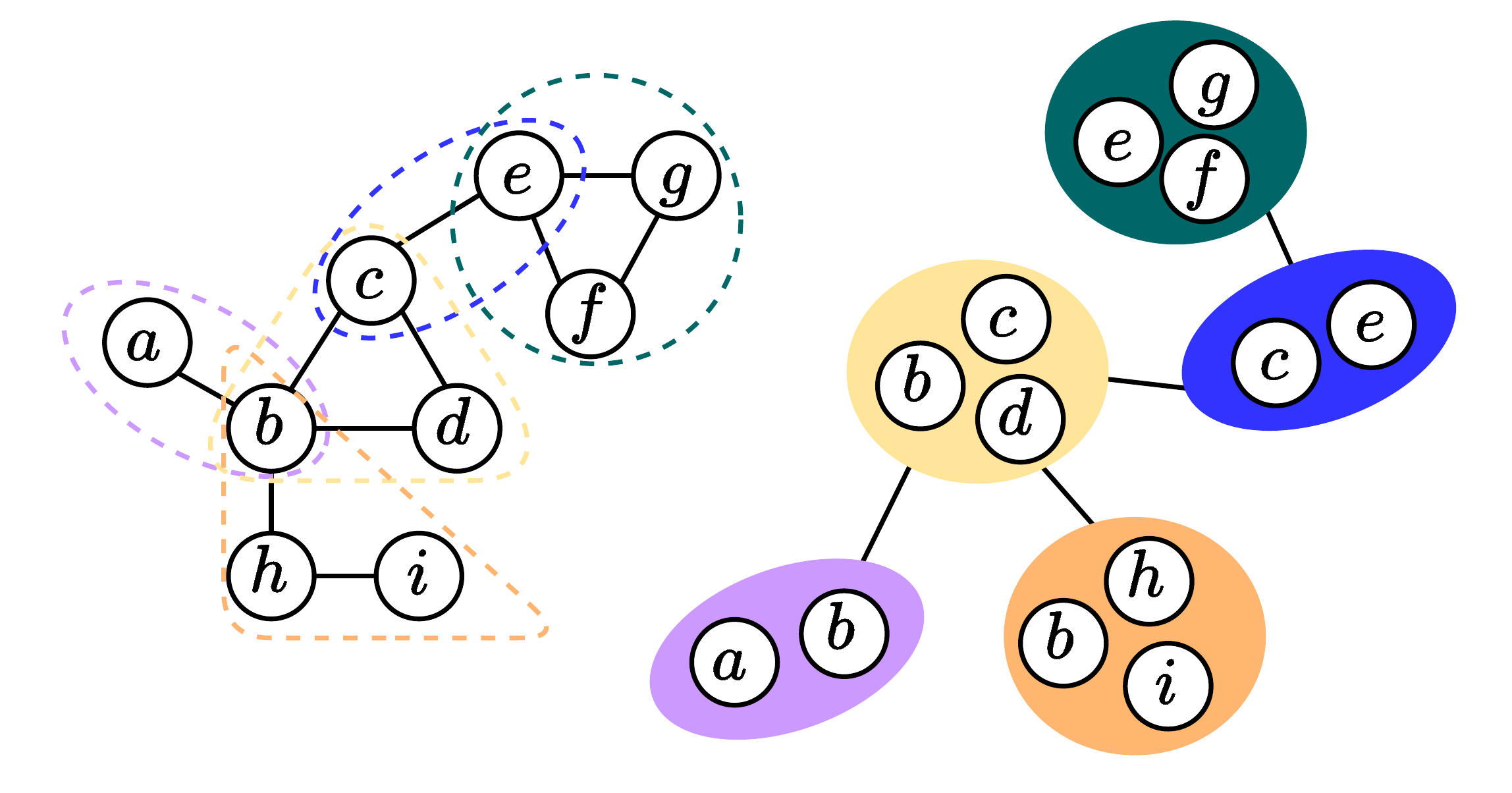}
  \caption{A graph and a width-two tree decomposition.
  Fixing the decisions on $\{b,e\}$ leaves at most two undecided vertices per
  bag, reducing the residual width to one. The disjoint bags $\{b,c,d\}$ and
  $\{e,f,g\}$ each require a fixed vertex, so two are necessary.}
  \label{fig:tm}
\end{figure}

We focus on dynamic programming (DP) parameterized by treewidth, which measures
how closely a graph's structure resembles a tree. A tree decomposition covers
the graph with overlapping vertex sets,
called \emph{bags}, arranged in a tree (Figure~\ref{fig:tm}). Every edge lies
in a bag, and the bags containing any given vertex form a connected
subtree~\citep{Robertson1984GraphMI}. The decomposition's width is the
\emph{largest bag size minus one}; treewidth is the minimum width over all
decompositions. The benefit of this DP is broad:
every fixed graph property expressible in monadic second-order logic can be
decided in linear time on graphs of bounded
treewidth~\citep{Courcelle1990TheMS}. In the dynamic programs used here,
the exponential dependence is on width, not graph
size~\citep{lokshtanov2011known}, so large graphs can be tractable when their
width is small.
Large width, however, still makes this computation expensive: the number of
possible local assignments grows exponentially with bag size.

We propose \textsc{N-FPT} to combine fast neural predictions with exact
completion, without retraining. On a fixed tree decomposition, it identifies
a treewidth modulator: a vertex set whose removal from the bags reduces the
width to a chosen threshold. A complete valid neural solution supplies its
decisions on this set as \emph{neural advice}; DP optimizes the remaining
decisions under the original constraints and objective.
We build on modulator-based enhancement of graph
heuristics~\citep{Li2026Taming}. While that framework repairs conflicting
advice when needed, our sequential neural samplers preserve feasibility at
each decision, supplying valid advice directly. We extend exact completion
to prediction consolidation and neural learning.

The neural solver therefore need not predict an entire optimal solution.
\textsc{N-FPT} returns and certifies the \emph{best achievable solution}
under its advice (Proposition~\ref{prop:nfpt}). It improves the neural
solution whenever a better compatible completion exists; otherwise, it
establishes that the solution is already optimal under the fixed advice.
Under \emph{perfect advice}---modulator decisions consistent with a global
optimum---the result is globally optimal. The framework is
\emph{neural-model agnostic}: the same procedure applies to any neural
solver supplying a valid solution.

However, perfect advice cannot be assumed in practice. We therefore make
improving advice quality part of the learning-augmented design, rather than
treating predictions as a fixed assumption. We develop \textsc{NIC-FPT} and
\textsc{NRD-FPT} as two compact \textsc{N-FPT} variants with
\emph{consolidated prediction}.
\textsc{NIC-FPT} commits selections supported by agreement across neural
samples and conditions subsequent samples on \emph{this consensus}.
\textsc{NRD-FPT} randomly defers selections and resamples while retaining
the others. Each variant returns the best achievable solution under its
consolidated modulator advice, without retraining the neural model.

Beyond inference, \textsc{N-FPT} offers a different learning paradigm:
learning to preserve optimal completion through exact corrections to sampled
decisions. A common approach is reinforcement learning, with rewards derived from
solution objective values~\citep{bello2016neural,khalil2017learning}.
This encourages high rewards on average, but a final score alone does not
reveal where optimal completion was lost. As a neural completion model (NC)
extends a partial solution, \textsc{N-FPT} supplies feedback once the
remainder becomes tractable; the original starting state need not admit
an affordable exact solve. From that point, its dynamic program represents
\emph{all optimal completions through shared local states}, without listing
complete solutions. NC continues making its own selections, while
\emph{\textsc{N-FPT} feedback} identifies the \emph{first subsequent decision
that rules out every optimal completion} and teaches alternatives that
preserve one. The stored states are reused across continuations from the
same partial solution. Neural decisions thus enable exact computation,
and exact feedback teaches NC to preserve the best value still achievable,
even when its samples miss that value.

Experiments across neural models and optimization problems demonstrate
solution-quality gains and robustness to graph-size and distribution
shifts. With matched architectures and starting states, feedback-trained NC
reaches optimal completions \emph{more often} than RL alone without an
exact solver at inference, so \emph{these gains reflect learned completion}.

Our contributions are:
\begin{itemize}[beginpenalty=10000]
  \item \emph{Best achievable solutions without neural retraining.}
        \textsc{N-FPT} accepts valid solutions from any neural model and
        certifies the best achievable completion of their modulator advice,
        with global optimality under perfect advice.
  \item \emph{Consolidated prediction within \textsc{N-FPT}.}
        \textsc{NIC-FPT} and \textsc{NRD-FPT} consolidate neural predictions
        through sample agreement and randomized deferral, respectively.
        Each returns the best achievable completion of the resulting advice
        without updating the neural model.
  \item \emph{Learning optimal completion from suboptimal samples.}
        As NC makes exact completion tractable, reusable \textsc{N-FPT}
        feedback identifies the first subsequent decision that loses
        optimal completion and teaches alternatives that preserve it,
        giving NC supervision beyond final rewards.
\end{itemize}

\section{Problem Setting} \label{sec:background}
We study maximum independent set (MIS), minimum vertex cover (MVC), and MaxCut
on $G=(V,E)$, with $n=|V|$ and $m=|E|$.
Let $\mathcal{F}(G)\subseteq 2^V$ be the valid solutions and $f_G(S)$ their
objective, maximized for MIS and MaxCut and minimized for MVC.
For a global optimum $S^\star$, write $\mathrm{OPT}(G)=f_G(S^\star)$.
Neural advice is a partial assignment $a:D\to\{0,1\}$ on
$D\subseteq V$, with $1$ meaning selected and $0$ excluded.
Its valid completions form $\mathcal{F}(G\mid a)$; we call $a$ valid when
this family is nonempty and write $V_G(a)$ for its \emph{best achievable value}.
An algorithm is \emph{fixed-parameter tractable} (FPT) in $k$ if it runs in
$f(k)\cdot n^{\mathcal{O}(1)}$ time, with computable $f$ and polynomial
exponent independent of $k$~\citep{cygan2015parameterized}.

\section{N-FPT: Exact Completion and Consolidated Prediction}
\label{sec:method}
\noindent\textbf{Making Exact Completion Tractable.}
We construct a tree decomposition
$\mathcal{T}=(T,\{X_t\}_{t\in V(T)})$ using the min-degree heuristic.
Here $X_t$ is the bag at node $t$ of the tree $T$.
For a target width $\eta\geq 0$, we compute a minimum modulator on this
fixed decomposition~\citep{Li2026Taming}:
\begin{equation}
  \mathcal{M}_{\eta}\in\arg\min_{M\subseteq V}|M|
  \quad\text{subject to}\quad
  |X_t\setminus M|\leq\eta+1
  \quad\text{for every }t\in V(T).
  \label{eq:fixed-decomposition-modulator}
\end{equation}
We solve this mixed-integer linear program exactly. The modulator is minimum
for the given decomposition, not over all decompositions of the graph.
Removing it from every bag gives a valid decomposition of
$G\setminus\mathcal{M}_{\eta}$ with width at most $\eta$
(Figure~\ref{fig:tm}).

Given a complete valid solution $S_\theta\in\mathcal{F}(G)$ from any neural
solver, we retain the advice $a_\theta(v)=\mathbf{1}\{v\in S_\theta\}$ for
$v\in\mathcal{M}_{\eta}$ and optimize all other decisions.
The neural model supplies only $S_\theta$; neither the modulator nor the exact
computation depends on its architecture or training objective.
The modulator vertices are fixed, not discarded: their constraints and
objective contributions remain part of the original problem. For MIS, for
example, a selected modulator vertex excludes its neighbors.
Advice from $S_\theta$ always admits a valid completion.

\begin{figure}[!ht]
  \centering
  \includegraphics[trim=0bp 12bp 0bp 8bp,clip,width=0.846\linewidth]{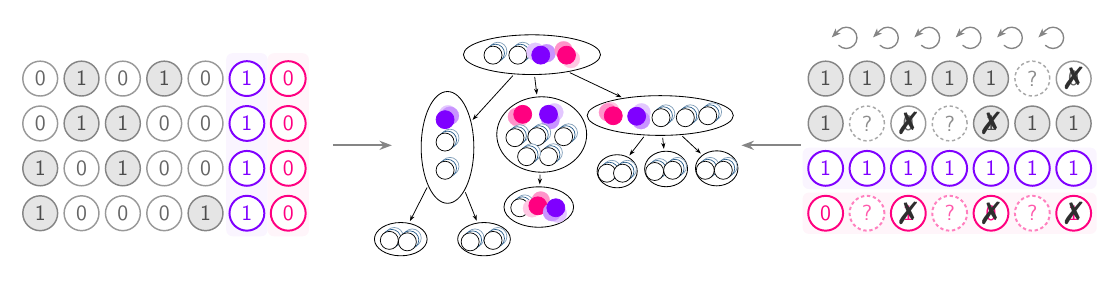}
  \caption{\textsc{N-FPT} (center): neural advice fixes colored vertices,
  and DP optimizes the remaining decisions on white vertices.
  \textsc{NIC-FPT} (left) and \textsc{NRD-FPT} (right) consolidate the advice;
  crosses indicate random deferral.}
  \label{fig:advice-refinement}
\end{figure}

The dynamic program retains the best compatible partial solution for each
of at most $2^{\eta+1}$ assignments to the undecided vertices in a bag.
Tracing back an optimal root entry returns a best achievable completion
(Figure~\ref{fig:advice-refinement}, center).
Given the decomposition and boundary contributions, completion is linear in
the number of bags and exponential in $\eta$. This FPT bound excludes
modulator optimization. Smaller $\eta$ reduces the exact computation but can
require more neural decisions; larger $\eta$ permits more decisions to be
optimized exactly. The relevant trade-off is thus residual width versus
dependence on neural advice, not graph size or density alone.

\begin{samepage}
\mdfsetup{nobreak=true}
\begin{proposition}[Exact completion, consistency, and non-degradation]
\label{prop:nfpt}
Fix a graph $G$, a modulator $\mathcal{M}_{\eta}$, and an exact
problem-specific treewidth dynamic program that enforces the advice and all
boundary constraints and objective contributions. Then:
\begin{itemize}[noitemsep]
  \item \emph{Exact completion.} For any valid advice
        $a:\mathcal{M}_{\eta}\to\{0,1\}$, \textsc{N-FPT} attains $V_G(a)$.
  \item \emph{Consistency.} If the advice is perfect, meaning that it is
        consistent with some globally optimal solution, the returned solution
        is globally optimal.
  \item \emph{Non-degradation.} For any complete valid neural solution
        $S_\theta$ and its restriction $a_\theta$ to $\mathcal{M}_{\eta}$,
        $V_G(a_\theta)\geq f_G(S_\theta)$ for maximization; the inequality
        is reversed for minimization.
\end{itemize}
\end{proposition}
\end{samepage}

The proof is in Appendix~\ref{app:consistency-proof}. For maximization, the
neural solution's gap separates as
\begin{equation}
  \mathrm{OPT}(G)-f_G(S_\theta)
  =\underbrace{\mathrm{OPT}(G)-V_G(a_\theta)}_{\text{advice regret}}
  +\underbrace{V_G(a_\theta)-f_G(S_\theta)}_{\text{loss in neural completion}}.
  \label{eq:advice-gap}
\end{equation}
For minimization, each difference is reversed. Exact completion removes the
second term; reducing the remaining advice regret requires better fixed
decisions. This motivates \textsc{NIC-FPT} and \textsc{NRD-FPT}.

\noindent\textbf{Consolidated Prediction.}
\textsc{NIC-FPT} and \textsc{NRD-FPT} consolidate neural advice within
\textsc{N-FPT} to reduce advice regret without retraining
(Figure~\ref{fig:advice-refinement}). Both require a neural sampler that
generates valid completions conditioned on fixed decisions.
For MIS, \textsc{NIC-FPT} draws $K$ valid completions per round, conditioned
on the consensus from earlier rounds. It fixes an undecided vertex as selected
when more than $\kappa$ samples select it, with $\kappa>K/2$, and excludes
its neighbors.
\textsc{NRD-FPT} instead starts from a complete valid neural sample and
independently defers each selected vertex with probability $\rho$ at each
round. Retained selections and their implied neighbor exclusions stay fixed
while the model resamples a valid completion over all other vertices.
Other problems use their own validity rules.

After the prescribed rounds, the final valid neural sample supplies
modulator advice $a$. Each variant returns its best achievable completion
$V_G(a)$, with the guarantees of Proposition~\ref{prop:nfpt}.

\begin{table}[!ht]
  \centering
  \setlength{\tabcolsep}{0.7ex}
  \renewcommand{\arraystretch}{0.9}
  \caption{Results across neural models on MIS (Reg graphs).
  Gaps use the starred Gurobi reference; improvements are compared within
  each neural model.}
  \label{tab:reg}

  {\scshape\small
    \begin{adjustbox}{max width=0.90\columnwidth}
      \begin{tabular}{ccccccccc}
        \toprule
        \multirow{2}{*}{Method}
          & \multicolumn{2}{c}{SS} & \multicolumn{2}{c}{SD} & \multicolumn{2}{c}{LS} & \multicolumn{2}{c}{LD} \\
        \cmidrule(lr){2-3}\cmidrule(lr){4-5}\cmidrule(lr){6-7}\cmidrule(lr){8-9}
          & Size $\uparrow$ & Gap\% $\downarrow$ & Size $\uparrow$ & Gap\% $\downarrow$
          & Size $\uparrow$ & Gap\% $\downarrow$ & Size $\uparrow$ & Gap\% $\downarrow$ \\
        \midrule
        \textsc{Gurobi(60s)} & 280.29 & 3.31 & 170.84 & 2.39 & 358.03 & 4.69 & 216.68 & 4.09 \\
        \textsc{*Gurobi(600s)} & 289.90 & 0.00 & 175.03 & 0.00 & 375.64 & 0.00 & 225.93 & 0.00 \\
        \textsc{KaMIS(600s)} & 299.97 & -3.47 & 183.96 & -5.10 & 387.11 & -3.05 & 238.28 & -5.47 \\
        \midrule
        \multicolumn{9}{l}{\textnormal{\textit{GFlowNet, average}}} \\
        \textsc{Standalone} & 286.46 & 1.19 & 170.40 & 2.65 & 372.55 & 0.82 & 221.34 & 2.03 \\
        \textsc{N-FPT} & 286.55 & 1.16 & 170.41 & 2.64 & 372.65 & 0.80 & \textbf{221.35} & \textbf{2.03} \\
        \textsc{NIC-FPT} & \textbf{286.86} & \textbf{1.05} & \textbf{171.71} & \textbf{1.90} & \textbf{372.88} & \textbf{0.73} & 221.34 & 2.03 \\
        \addlinespace[3pt]
        \multicolumn{9}{l}{\textnormal{\textit{GFlowNet, best-of-20}}} \\
        \textsc{Standalone} & 290.55 & -0.22 & 174.67 & 0.21 & 377.22 & -0.42 & 226.23 & -0.13 \\
        \textsc{N-FPT} & 290.63 & -0.25 & 174.68 & 0.20 & 377.35 & -0.46 & 226.25 & -0.14 \\
        \textsc{NRD-FPT} & \textbf{293.90} & \textbf{-1.38} & \textbf{176.96} & \textbf{-1.10} & \textbf{381.78} & \textbf{-1.63} & \textbf{228.68} & \textbf{-1.22} \\
        \midrule
        \multicolumn{9}{l}{\textnormal{\textit{GNN+RL, average}}} \\
        \textsc{Standalone} & 290.63 & -0.25 & 174.32 & 0.41 & 374.06 & 0.42 & 225.46 & 0.21 \\
        \textsc{N-FPT} & 290.66 & -0.26 & 174.32 & 0.41 & 374.09 & 0.41 & 225.46 & 0.21 \\
        \textsc{NIC-FPT} & \textbf{290.84} & \textbf{-0.32} & \textbf{174.33} & \textbf{0.40} & \textbf{374.20} & \textbf{0.38} & \textbf{225.47} & \textbf{0.20} \\
        \addlinespace[3pt]
        \multicolumn{9}{l}{\textnormal{\textit{GNN+RL, best-of-20}}} \\
        \textsc{Standalone} & 294.48 & -1.58 & 177.92 & -1.65 & 378.51 & -0.76 & 230.02 & -1.81 \\
        \textsc{N-FPT} & 294.51 & -1.59 & 177.92 & -1.65 & 378.54 & -0.77 & 230.02 & -1.81 \\
        \textsc{NRD-FPT} & \textbf{296.70} & \textbf{-2.35} & \textbf{180.25} & \textbf{-2.98} & \textbf{381.88} & \textbf{-1.66} & \textbf{233.07} & \textbf{-3.16} \\
        \bottomrule
      \end{tabular}
    \end{adjustbox}
  }
\end{table}

\noindent\textbf{Experimental Overview.}
We test whether \textsc{N-FPT} strengthens neural solutions across
\emph{neural models}, \emph{optimization problems}, and
\emph{test distributions}. Each trained solver
is compared with its \textsc{N-FPT}, \textsc{NIC-FPT}, and \textsc{NRD-FPT}
versions on the same test instances, without updating its weights.
The objective is improvement to each existing neural solver; Gurobi and
KaMIS provide reference values. Tables~\ref{tab:reg}
and~\ref{tab:cross-problem} report random regular (Reg) graphs in four
settings combining small/large graphs (S/L) with sparse/dense settings
(S/D). Full protocols and additional graph families are
provided in the appendix.

\noindent\textbf{Results across Neural Models.}
We first compare two established learning approaches on MIS
(Table~\ref{tab:reg}). GFlowNet learns a reward-weighted distribution over
solutions~\citep{zhang2023letflowstellsolving}. GNN+RL represents the widely
used combination of graph neural networks and reinforcement
learning~\citep{khalil2017learning,barrett2020exploratory,Ahn2020LearningWT};
our policy maximizes expected solution reward. \emph{Exact completion
complements both learning objectives}: the same parameterized procedure
recovers improvements left unused by either trained solver.

A fractional increase in mean MIS size represents whole-vertex improvements
in individual solutions, averaged over graphs, samples, and training seeds.
\textsc{N-FPT} reaches the best value permitted by each solution's advice,
regardless of how well the neural model is trained. If exact completion
leaves an individual solution's objective unchanged, it \emph{certifies
that solution as best achievable under its fixed advice}.
Once this value is reached, further gains require better advice.
The additional gains from \textsc{NIC-FPT} and \textsc{NRD-FPT} demonstrate
the value of consolidated prediction.

\noindent\textbf{Results across Optimization Problems.}
We next apply the framework to MaxCut and MVC using GFlowNet
(Table~\ref{tab:cross-problem}). MaxCut tests an edge-based maximization
objective, while MVC tests vertex-based minimization.
\textsc{N-FPT} improves both objectives, while consolidated prediction
enables further gains; \textsc{NRD-FPT} improves best-of-20 quality in every
reported setting for both problems. Together with the cross-model results,
these gains demonstrate applicability across \emph{neural models and
optimization problems}.

\begin{table}[!t]
  \centering
  \setlength{\tabcolsep}{0.0ex}
  \renewcommand{\arraystretch}{0.9}
  \caption{Results across optimization problems with GFlowNet on Reg graphs.
  Mean objective values.}
  \label{tab:cross-problem}

  {\scshape\footnotesize
    \begin{adjustbox}{max width=\linewidth}
      \begin{tabular}{@{}l*{4}{@{\hspace{0.6ex}}c}@{}p{2.4ex}@{}l*{4}{@{\hspace{0.6ex}}c}@{}}
        \multicolumn{5}{c}{\textnormal{\textbf{MaxCut $\uparrow$}}}
          & & \multicolumn{5}{c}{\textnormal{\textbf{MVC $\downarrow$}}} \\[2pt]
        \cmidrule[\heavyrulewidth]{1-5}\cmidrule[\heavyrulewidth]{7-11}
        Method & \makebox[3.8em]{SS} & \makebox[3.8em]{SD} & \makebox[3.8em]{LS} & \makebox[3.8em]{LD} & & Method & \makebox[3.8em]{SS} & \makebox[3.8em]{SD} & \makebox[3.8em]{LS} & \makebox[3.8em]{LD} \\
        \cmidrule{1-5}\cmidrule{7-11}
        \multicolumn{5}{l}{\textnormal{\textit{Average}}}
          & & \multicolumn{5}{l}{\textnormal{\textit{Average}}} \\[-1pt]
        \textsc{Standalone} & 1944.50 & 4458.06 & 2582.00 & 5767.26 &  & \textsc{Standalone} & 561.59 & 680.14 & 729.77 & 881.69 \\
        \textsc{N-FPT} \textnormal{(ours)} & 1967.72 & 4474.58 & 2587.64 & 5792.78 &  & \textsc{N-FPT} \textnormal{(ours)} & 561.50 & 680.13 & 729.67 & \textbf{881.68} \\
        \textsc{NIC-FPT} \textnormal{(ours)} & \textbf{1996.30} & \textbf{4490.85} & \textbf{2608.50} & \textbf{5824.44} &  & \textsc{NIC-FPT} \textnormal{(ours)} & \textbf{561.19} & \textbf{680.07} & \textbf{729.44} & 881.69 \\
        \addlinespace[2pt]
        \multicolumn{5}{l}{\textnormal{\textit{Best-of-20}}}
          & & \multicolumn{5}{l}{\textnormal{\textit{Best-of-20}}} \\[-1pt]
        \textsc{Standalone} & 1963.73 & 4492.65 & 2601.64 & 5804.41 &  & \textsc{Standalone} & 557.50 & 675.87 & 725.10 & 876.80 \\
        \textsc{N-FPT} \textnormal{(ours)} & 1986.51 & 4509.91 & 2606.80 & 5830.97 &  & \textsc{N-FPT} \textnormal{(ours)} & 557.42 & 675.86 & 724.97 & 876.78 \\
        \textsc{NRD-FPT} \textnormal{(ours)} & \textbf{2007.57} & \textbf{4544.66} & \textbf{2618.95} & \textbf{5894.62} &  & \textsc{NRD-FPT} \textnormal{(ours)} & \textbf{554.15} & \textbf{673.58} & \textbf{720.54} & \textbf{874.35} \\
        \cmidrule[\heavyrulewidth]{1-5}\cmidrule[\heavyrulewidth]{7-11}
      \end{tabular}
    \end{adjustbox}
  }
\end{table}

\begin{figure}[!ht]
  \centering
  \includegraphics[width=0.846\linewidth]{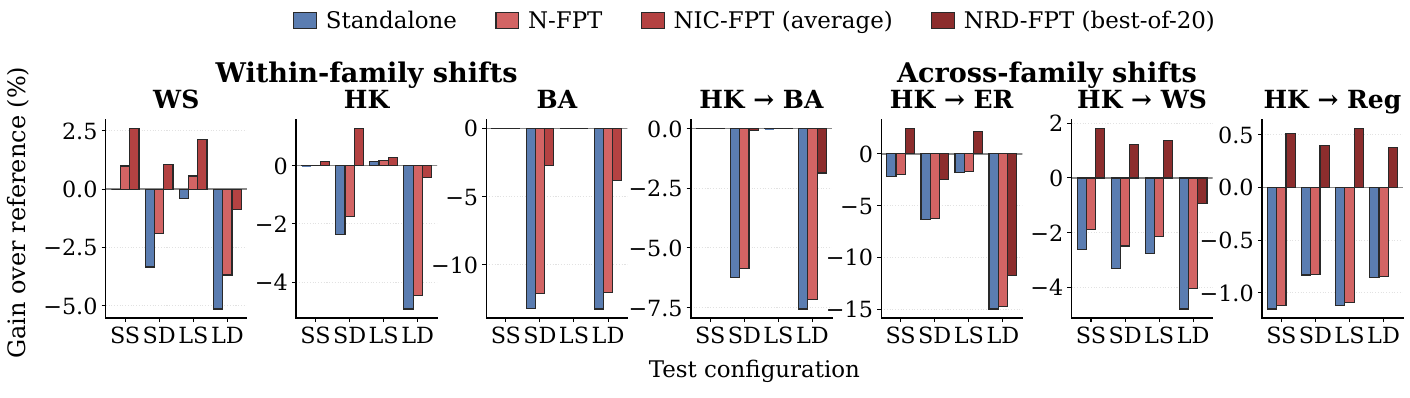}
  \caption{Gains in average solution size (left) and best-of-20 solution
  size (right).
  All methods and the reference use the same measure within each block.}
  \label{fig:generalization}
\end{figure}

\FloatBarrier
\noindent\textbf{Generalization across Graph Sizes and Distributions.}
We evaluate GFlowNet trained on small sparse (SS) graphs for MIS across
graph sizes, densities, and families, without retraining
(Figure~\ref{fig:generalization}). Zero marks the performance of a standalone GFlowNet
trained specifically for the target configuration; positive values exceed
this reference, which is not a certified optimum.

Changing density can be more disruptive than increasing graph size alone.
Across graph families, the same trained solver can also lose quality.
\textsc{N-FPT} improves these transferred solutions, while \textsc{NIC-FPT}
and \textsc{NRD-FPT} reduce the remaining losses and sometimes surpass the
target-trained reference. Thus, \emph{training on the test distribution is
not required to benefit from the framework}.

\FloatBarrier

\section{Learning Optimal Completion with N-FPT Feedback}
\label{sec:learning}
\begin{figure}[!t]
  \centering
  \includegraphics[width=0.719\linewidth,keepaspectratio]{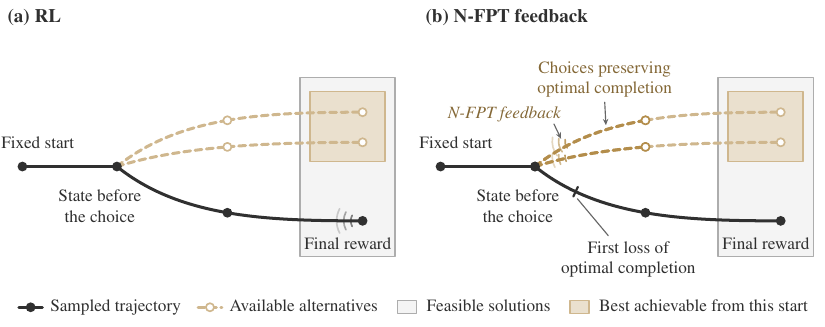}
  \caption{Learning signals from a partial state where exact completion is tractable.
  Dashed paths are available alternatives, not additional samples;
  gold waves mark \textsc{N-FPT} feedback.}
  \label{fig:nc-trajectory-feedback}
\end{figure}

Training a neural solver for high solution rewards does not guarantee an
optimal solution. In our RL baseline, a final reward neither locates the
decision that lost optimal completion nor identifies alternatives that would
have preserved it (Figure~\ref{fig:nc-trajectory-feedback}(a)).
Obtaining optimal training targets for NP-hard problems can itself exceed
practical budgets on large, difficult instances: an exact solver must both
find an optimum and prove that no better solution exists.
With no initial assignment, the target is \emph{global optimality}; once
fixed decisions rule out every global optimum, it is the \emph{optimal
completion conditioned on those decisions}. The missing supervision is
\textbf{which decisions preserve the best achievable completion}, not just
final solution quality.

\emph{\textsc{N-FPT} feedback} supplies this supervision by retaining the
dynamic program's optimal alternatives, not just one completed solution.
As NC proceeds along a trajectory, feedback becomes available when the
remaining problem becomes tractable. NC continues making selections, while
these alternatives identify the first subsequent selection that loses the
best achievable value and the choices that preserve it
(Figure~\ref{fig:nc-trajectory-feedback}(b)). Feedback at the state before
that selection teaches NC to preserve optimal completion on future
trajectories; sampled choices remain unchanged. Neural progress enables
exact feedback without requiring the original starting state to be
exactly solvable within the computation budget.

\noindent\textbf{Learning to Preserve Optimal Completion.}
We describe the construction for MIS. Let $a_0$ be a valid, nonterminal
starting assignment.
NC's policy $\pi_\theta(v\mid G,a)$ observes the full graph and all decisions
made so far, including $a_0$. Each vertex is selected,
excluded, or undecided. A move selects an undecided vertex and excludes its
neighbors, producing the next assignment; selections continue until no vertex
is undecided. Fixed decisions are never changed.
Write $a_t$ for the assignment after $t$ moves. Let $s\geq0$ denote the step
at which a nonterminal remainder becomes tractable. At $a_s$, \textsc{N-FPT} computes
$V_G(a_s)$ and activates feedback for the continuation. All decisions through
$a_s$ stay fixed; feedback concerns the best value achievable from that point.
The optimal completions still available are
\begin{equation}
  \mathcal{O}_t
  =\{S\in\mathcal{F}(G\mid a_t): f_G(S)=V_G(a_s)\},\qquad t\geq s.
  \label{eq:surviving-optima}
\end{equation}
Because each move fixes further decisions, these families are nested:
$\mathcal{O}_t\subseteq\mathcal{O}_{t-1}$ for $t>s$.
A continuation that ends below $V_G(a_s)$ must therefore have a
\emph{first harmful selection} after feedback becomes available, at
$h=\min\{t>s:\mathcal{O}_t=\varnothing\}$.
Before this selection, the best achievable value is still $V_G(a_s)$;
after it, that value cannot be recovered without changing a fixed decision.
Feedback at $a_{h-1}$ identifies all next selections that preserve at least
one member of $\mathcal{O}_{h-1}$. It accepts every such choice, without
prescribing one optimal solution.

\noindent\textbf{Reusable Feedback from the Dynamic Program.}
A single optimal solution from an exact solver such as Gurobi~\citep{gurobi}
is not enough to label every neural decision: a different choice may still
belong to another optimum. Exhaustive enumeration can recover all optimal
alternatives, but storing them individually can require exponential space
even at treewidth one. For example, $m$ disjoint edges admit $2^m$ maximum
independent sets, while their dynamic program needs only two optimal local
choices per edge. We exploit this sharing to retain \emph{all optimal
completions without listing them}. For fixed residual width, the shared
representation uses space polynomial in the decomposition size, even when
the number of complete optima is exponential.

At $a_s$, we extend the exact completion procedure of \textsc{N-FPT} to
retain optimal alternatives rather than a single traceback.
Starting from all optimal root entries, we keep every child entry that
agrees with its parent on shared vertices and attains the best subtree value
under that shared assignment. The retained bag states and their compatibility
links represent all of $\mathcal{O}_s$ without listing complete solutions.
We build this representation once for $a_s$ and reuse it whenever NC
continues from that same partial assignment.

As NC continues beyond $a_s$, we follow its selections through the stored
representation. Selecting a vertex and excluding its neighbors deactivates
inconsistent bag states. A parent state is also deactivated if it loses all
compatible states in any child bag, propagating the change toward the root.
An optimal completion survives exactly when at least one root state remains
active. The first selection that removes the last active root state is
therefore the first harmful selection after $a_s$. We undo that update only
in the stored representation and follow compatible active entries from the surviving roots:
every undecided vertex occurring in these entries belongs to a remaining
optimal completion and is an acceptable next selection. Root connectivity
ensures that the feedback refers to a complete optimal solution, not merely
a locally compatible bag assignment. Each continuation from $a_s$ resets
only the activity flags and support counts. Once the representation is
available, tracking its selections requires \emph{no further exact solves}.

\noindent\textbf{Training RL with N-FPT Feedback.}
We use this feedback to train NC with a single policy objective that rewards
good solutions while teaching decisions that preserve optimal completion.
From the same starting assignment $a_0$, NC samples $K$ complete trajectories
from its own policy. Each trajectory is a sequence of valid selections that
finishes the solution, without being forced to follow an optimal completion.
As a trajectory reaches a tractable state $a_s$, exact feedback becomes
available for its subsequent decisions; different trajectories can reach
different such states. Our experiments use $K=4$.

For each state $a_s$ where feedback is available, a feedback example $(a,B)$
consists of a partial assignment $a$ at or after $a_s$ and a set $B$ of next
selections known to preserve $V_G(a_s)$.
We obtain these pairs along a randomly ordered exact completion of $a_s$
and immediately before the first harmful selection in NC's continuation.
Along the exact completion, $B$ contains its remaining selected
vertices; at a sampled mistake, $B$ contains all next selections supported
by the surviving optimal completions. Both use the same feedback rule:
increase the policy's total probability on $B$. No particular order is
imposed on these choices, and the sampled trajectories remain unchanged.

For trajectory $i$, let $L_i$ be the number of additional selections made
after $a_0$; decisions already fixed in $a_0$ do not count toward $L_i$.
At step $t$, NC selects vertex $v_{i,t}$ from the current assignment
$a_{i,t-1}$, producing $a_{i,t}$. All trajectories begin at $a_{i,0}=a_0$.
The final solution $S_i$ receives reward $r_i=f_G(S_i)$.
Its advantage $A_i$ is $r_i$ minus the mean reward of the other $K-1$
trajectories from the same start. This comparison requires at least two
trajectories ($K\geq2$).

Let $q_{\mathrm{fb}}$ sample feedback pairs uniformly within each of the two
sources above, giving each source equal total weight when both are available.
We minimize the combined loss
\begin{equation}
  \mathcal{L}_{\mathrm{NC}}(\theta)
  =-\frac{1}{K}\sum_{i=1}^{K} A_i
    \sum_{t=1}^{L_i}\log\pi_\theta(v_{i,t}\mid G,a_{i,t-1})
   -\lambda\,\mathbb{E}_{(a,B)\sim q_{\mathrm{fb}}}
    \log\!\left(\sum_{v\in B}\pi_\theta(v\mid G,a)\right),
  \label{eq:nfpt-feedback-loss}
\end{equation}
\begin{figure}[!t]
  \centering
  \includegraphics[width=0.762\linewidth]{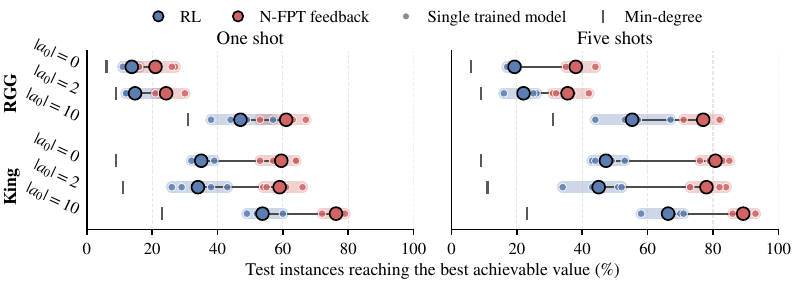}
  \caption{Optimal-completion frequency. Dots show four-seed means (large)
  and individual models (small); pale bars span seed ranges and gray ticks
  show min-degree. $|a_0|$ counts initial selections.}
  \label{fig:nc-optimal-completion}
\end{figure}
where $\lambda\geq0$ controls the feedback weight. Recorded trajectories,
advantages, and feedback pairs are held fixed during the update.
The feedback term is omitted when no feedback examples are available.
The policy is normalized over all undecided vertices, so feedback changes
what NC learns without restricting which valid trajectories it can sample.
Even when none of those trajectories is optimal, the exact computation
supplies decisions that would preserve the best achievable value.

The trained policy can complete a new partial solution on its own or supply
further decisions before \textsc{N-FPT} finishes a tractable remainder.
The feedback is exact; its effect on learned completion is evaluated below.

\noindent\textbf{Experimental Overview.}
We compare \emph{RL} and \emph{RL with \textsc{N-FPT} feedback} on MIS,
using matched architectures, training graphs, starting assignments, and
optimizer-update counts. Tests use random geometric graphs (RGG) and king's
graphs with random vertex deletions (King), with zero, two, or ten initial
selections. NC completes each test solution without an exact solver and is
assessed against its starting assignment's proven optimum.
\emph{One shot} selects the highest-probability vertex at each step;
\emph{five shots} returns the best of five sampled completions.
These runs use $s=0$ wherever exact feedback is available;
the appendix gives the full protocol and additional five-shot results.

\begin{figure}[!t]
  \centering
  \includegraphics[width=0.846\linewidth]{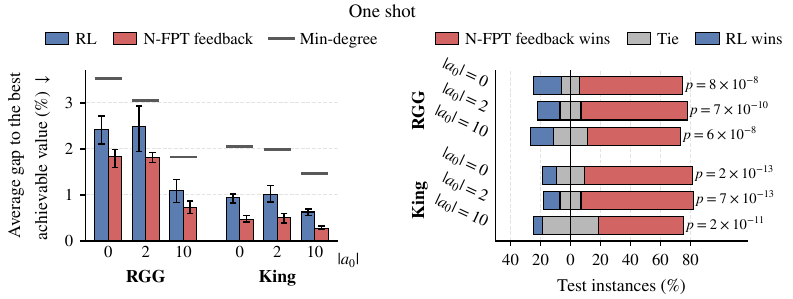}
  \caption{One-shot completion quality. Left: four-seed mean gaps with
  min--max whiskers; gray dashes show min-degree. Right: wins, ties, and losses
  based on four-seed mean solution sizes, with ties centered on zero.
  The $p$-values test equal win probabilities, excluding ties.}
  \label{fig:nc-completion-quality}
\end{figure}

\noindent\textbf{Reaching Optimal Completions More Often.}
Figure~\ref{fig:nc-optimal-completion} shows a higher mean optimal-completion
rate with \textsc{N-FPT} feedback in all twelve comparisons, with gains of
up to 33.5 percentage points. The gains already appear in one-shot decoding,
so they do not depend on additional test-time sampling. Under five-shot
decoding, even the weakest of the four feedback-trained models outperforms
the strongest RL baseline model in every setting. The improvement therefore
extends beyond a favorable training seed.

With $|a_0|=0$, reaching the best achievable value means finding a
\emph{global optimum}; for nonempty starts, it means finding the optimum
compatible with the fixed decisions. Across both graph families,
feedback-trained NC reaches this target more often and has smaller average
gaps from ten-selection starts than from empty or two-selection starts
(Figures~\ref{fig:nc-optimal-completion} and~\ref{fig:nc-completion-quality}).
These larger starts leave fewer decisions for NC, while feedback improves
over RL at every starting size. Its gains therefore cover both solving from
scratch and completing partial solutions without revising their fixed decisions.

\noindent\textbf{Solution Quality across Test Instances.}
Feedback lowers the average gap left by an RL baseline that already
outperforms min-degree (Figure~\ref{fig:nc-completion-quality}, left).
The paired comparisons (right) show that it improves many more test instances
than it worsens. The benefit is therefore not confined to a few favorable
completions: NC produces better solutions across the test set, as well as
reaching the best achievable value more often.

\begin{figure}[!ht]
  \centering
  \includegraphics[width=0.592\linewidth]{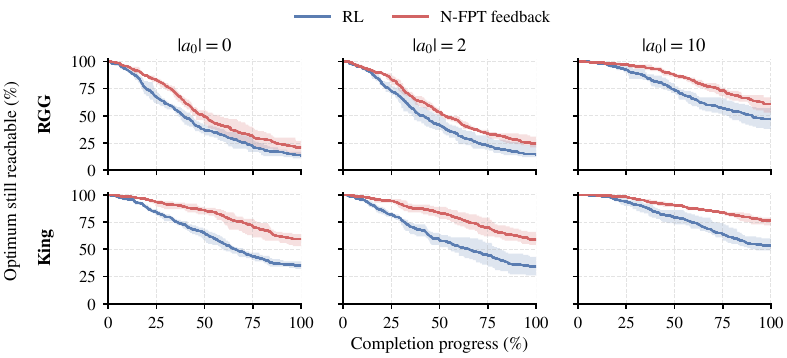}
  \caption{Optimal completion along one-shot test trajectories. Progress is
  the fraction of each trajectory's additional selections. Lines show
  four-seed means; bands span seed ranges.}
  \label{fig:nc-optimum-survival}
\end{figure}

\noindent\textbf{Preserving Optimal Completion along Trajectories.}
On one-shot test trajectories, \textsc{N-FPT} feedback keeps an optimal
completion possible on more graphs as the policy makes its selections
(Figure~\ref{fig:nc-optimum-survival}). The endpoints reproduce the one-shot
optimal-completion rates in Figure~\ref{fig:nc-optimal-completion}.
This connects the solution-quality gains to \emph{preserving optimal
completion during decoding}, the behavior that the feedback explicitly
teaches.

\FloatBarrier

\begin{figure}[!ht]
  \centering
  \includegraphics[width=0.592\linewidth]{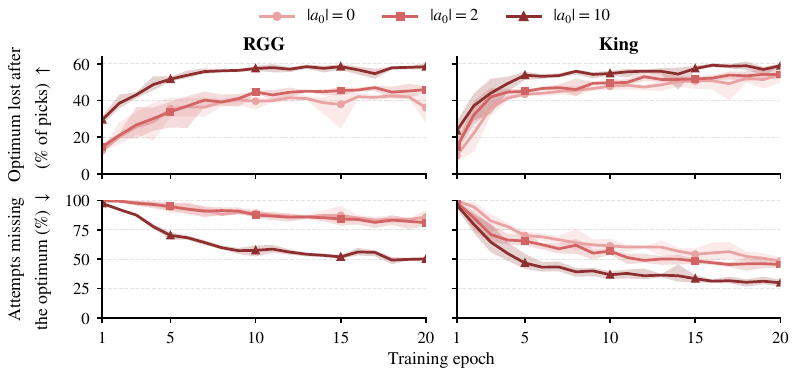}
  \caption{Training with \textsc{N-FPT} feedback. Top: the first harmful selection
  as a percentage of each suboptimal trajectory's additional selections,
  averaged over those trajectories. Lines show four-seed means; bands span
  seed ranges.}
  \label{fig:nc-feedback-training}
\end{figure}

Figure~\ref{fig:nc-feedback-training} shows that \emph{suboptimal samples can
teach optimal completion}. Early in training, nearly all samples miss the
best achievable value, yet \textsc{N-FPT} provides exact corrections.
Across both graph families and all starting sizes, fewer trajectories lose
optimal completion over training; among remaining failures, the first
harmful selection generally occurs later. Read together, these trends
support the intended learning behavior: NC improves both its chance of
reaching the optimum and its ability to preserve that possibility through
successive decisions.

\FloatBarrier

\section{Conclusion}
\textsc{N-FPT} uses parameterized algorithms to improve neural solutions
and guide learning for graph combinatorial optimization. The
neural-model-agnostic framework returns and certifies the \emph{best
achievable solution} under valid neural advice without retraining, and
recovers a global optimum under perfect advice.
Its \textsc{NIC-FPT} and \textsc{NRD-FPT} variants improve advice through
consolidated prediction. As NC makes exact completion tractable,
\textsc{N-FPT} feedback teaches which subsequent decisions preserve
optimal completion, even when sampled solutions are suboptimal.
Experiments show solution-quality gains across neural
models and optimization problems, with robustness to graph-size and
distribution shifts. Feedback-trained models also reach optimal completions
more often than RL alone, without an exact solver at inference.
Parameterized algorithms thus provide both \emph{guarantees for completing
neural advice} and \emph{precise guidance for learning better decisions}.

\bibliography{references}
\bibliographystyle{plainnat}

\appendix

\section{Proof and Further Results}

Algorithm~\ref{alg:tdpa} gives the exact completion procedure in \textsc{N-FPT}. Its state validity
and transition rules are instantiated for each optimization problem.
\begin{algorithm*}[!ht]
  \caption{Exact completion in \textsc{N-FPT}}
  \label{alg:tdpa}
  \begin{algorithmic}[1]
    \State \textbf{Input:} Graph $G=(V,E)$; rooted tree decomposition
    $\mathcal{T}=(T,\{X_t\}_{t\in V(T)})$; modulator
    $\mathcal{M}_{\eta}$; valid advice
    $a:\mathcal{M}_{\eta}\to\{0,1\}$; vertex-selection problem $P$.
    \State \textbf{Output:} The best achievable solution consistent with $a$.
    \For{$t\in V(T)$ in bottom-up order}
      \State $U_t\gets X_t\setminus\mathcal{M}_{\eta}$
      \State Enumerate assignments $z\in\{0,1\}^{U_t}$ and combine each with
      $a$ restricted to $X_t\cap\mathcal{M}_{\eta}$.
      \State Discard locally invalid assignments and assignments incompatible
      with the retained entries of a child bag.
      \State Apply the problem-specific DP transition and retain the best entry
      for each assignment on the intersection with the parent bag.
    \EndFor
    \Return the best entry at the root and its reconstructed solution.
  \end{algorithmic}
\end{algorithm*}

\subsection{Proof of Proposition~\ref{prop:nfpt}}
\label{app:consistency-proof}

\begin{proof}
Fix any valid advice $a$. By definition, $\mathcal{F}(G\mid a)$ contains
exactly the valid complete solutions that agree with $a$ on
$\mathcal{M}_{\eta}$. The dynamic program fixes those advised decisions and enumerates every
remaining bag assignment permitted by the problem-specific DP. The standard
tree-decomposition transitions discard exactly the invalid or mutually
inconsistent assignments. Because the dynamic program is exact, its reconstructed output is
an optimizer over $\mathcal{F}(G\mid a)$. Its objective value is therefore
$V_G(a)$, proving exact completion.

For consistency, suppose that a global optimum $S^\star$ agrees with $a$ on
$\mathcal{M}_{\eta}$. Then $S^\star\in\mathcal{F}(G\mid a)$. No solution in
this restricted set can be better than a global optimum, while exact completion
returns its best member. Hence the solution returned by \textsc{N-FPT} is globally optimal.

For non-degradation, $a_\theta$ is the restriction of the complete valid neural
solution $S_\theta$ to $\mathcal{M}_{\eta}$. Thus
$S_\theta\in\mathcal{F}(G\mid a_\theta)$. Exact optimization over a set that
contains $S_\theta$ cannot return a smaller value for maximization or a larger
value for minimization. The inequality is strict precisely when
$\mathcal{F}(G\mid a_\theta)$ contains a solution better than $S_\theta$.
\end{proof}

\subsection{More Results on Solution Quality}
\label{app:solution-quality}
These tables supplement the cross-model and cross-problem comparisons in
Section~\ref{sec:method} with further graph families and reference values.
Dataset specifications and the evaluation protocol are
collected in Appendices~\ref{appendix:dataset}
and~\ref{app:inference-evaluation}.

\paragraph{Table conventions.}
\label{app:inference-table-conventions}
The following solution-quality tables use GFlowNet. \textsc{Standalone}
denotes the neural solver alone; \textsc{NIC-FPT} and \textsc{NRD-FPT} are
the consolidated-prediction variants of \textsc{N-FPT}.
Average results compare \textsc{Standalone}, \textsc{N-FPT}, and
\textsc{NIC-FPT}; best-of-20 results compare \textsc{Standalone},
\textsc{N-FPT}, and \textsc{NRD-FPT}. Both groups follow the
aggregation described in Appendix~\ref{app:inference-evaluation}.

Gaps are percentages relative to the starred Gurobi (600s) reference mean
$b$, not certified optimality gaps. For reported mean size $x$, the gap is
$100(1-x/b)$ for MIS and MaxCut, and $100(x/b-1)$ for MVC. Negative gaps
indicate a better value than this reference. Within each sampling group,
bold marks the best reported neural value in each column, including ties
at the displayed precision. Table~\ref{tab:reg-ablations} reports the
GFlowNet results from Table~\ref{tab:reg}.
Tables~\ref{tab:mcut-reg} and~\ref{tab:mvc-reg} give the full results behind
Table~\ref{tab:cross-problem}, including solver references and reference gaps.

\begin{table}[!ht]
  \centering
  \setlength{\tabcolsep}{0.0ex}
  \renewcommand{\arraystretch}{0.9}
  \caption{GFlowNet results on Reg graphs for MIS. Cross-model results
  appear in Table~\ref{tab:reg}; reporting conventions are in
  Appendix~\ref{app:inference-table-conventions}.}
  \label{tab:reg-ablations}

  {\scshape\footnotesize
    \begin{adjustbox}{max width=\columnwidth}
      \begin{tabular}{@{}l*{4}{@{\hspace{1.5ex}}c@{\hspace{1.0ex}}c}@{}}
        \toprule
        \multirow{2}{*}{Method}
          & \multicolumn{2}{c}{SS} & \multicolumn{2}{c}{SD}
          & \multicolumn{2}{c}{LS} & \multicolumn{2}{c}{LD} \\
        \cmidrule(lr){2-3}\cmidrule(lr){4-5}\cmidrule(lr){6-7}\cmidrule(lr){8-9}
          & Size $\uparrow$ & Gap\% $\downarrow$ & Size $\uparrow$ & Gap\% $\downarrow$
          & Size $\uparrow$ & Gap\% $\downarrow$ & Size $\uparrow$ & Gap\% $\downarrow$ \\
        \midrule
        \textsc{Gurobi} \textnormal{(60s)} & 280.29 & 3.31 & 170.84 & 2.39 & 358.03 & 4.69 & 216.68 & 4.09 \\
        \textsc{Gurobi} \textnormal{(600s)}$^*$ & 289.90 & 0.00 & 175.03 & 0.00 & 375.64 & 0.00 & 225.93 & 0.00 \\
        \midrule
        \multicolumn{9}{l}{\textnormal{\textit{GFlowNet, average}}} \\[-1pt]
        \textsc{Standalone} & 286.46 & 1.19 & 170.40 & 2.65 & 372.55 & 0.82 & 221.34 & \textbf{2.03} \\
        \textsc{N-FPT} \textnormal{(ours)} & 286.55 & 1.16 & 170.41 & 2.64 & 372.65 & 0.80 & \textbf{221.35} & \textbf{2.03} \\
        \textsc{NIC-FPT} \textnormal{(ours)} & \textbf{286.86} & \textbf{1.05} & \textbf{171.71} & \textbf{1.90} & \textbf{372.88} & \textbf{0.73} & 221.34 & \textbf{2.03} \\
        \addlinespace[2pt]
        \multicolumn{9}{l}{\textnormal{\textit{GFlowNet, best-of-20}}} \\[-1pt]
        \textsc{Standalone} & 290.55 & -0.22 & 174.67 & 0.21 & 377.22 & -0.42 & 226.23 & -0.13 \\
        \textsc{N-FPT} \textnormal{(ours)} & 290.63 & -0.25 & 174.68 & 0.20 & 377.35 & -0.46 & 226.25 & -0.14 \\
        \textsc{NRD-FPT} \textnormal{(ours)} & \textbf{293.90} & \textbf{-1.38} & \textbf{176.96} & \textbf{-1.10} & \textbf{381.78} & \textbf{-1.63} & \textbf{228.68} & \textbf{-1.22} \\
        \bottomrule
      \end{tabular}
    \end{adjustbox}
  }
\end{table}

\begin{table}[!ht]
  \centering
  \setlength{\tabcolsep}{0.0ex}
  \renewcommand{\arraystretch}{0.9}
  \caption{MIS on WS graphs with GFlowNet. Reporting conventions are in
  Appendix~\ref{app:inference-table-conventions}.}
  \label{tab:ws}

  {\scshape\footnotesize
    \begin{adjustbox}{max width=\columnwidth}
      \begin{tabular}{@{}l*{4}{@{\hspace{1.5ex}}c@{\hspace{1.0ex}}c}@{}}
        \toprule
        \multirow{2}{*}{Method}
          & \multicolumn{2}{c}{SS} & \multicolumn{2}{c}{SD}
          & \multicolumn{2}{c}{LS} & \multicolumn{2}{c}{LD} \\
        \cmidrule(lr){2-3}\cmidrule(lr){4-5}\cmidrule(lr){6-7}\cmidrule(lr){8-9}
          & Size $\uparrow$ & Gap\% $\downarrow$ & Size $\uparrow$ & Gap\% $\downarrow$
          & Size $\uparrow$ & Gap\% $\downarrow$ & Size $\uparrow$ & Gap\% $\downarrow$ \\
        \midrule
        \textsc{Gurobi} \textnormal{(60s)} & 133.79 & 0.00 & 94.43 & 0.02 & 197.74 & 0.01 & 138.64 & 0.23 \\
        \textsc{Gurobi} \textnormal{(600s)}$^*$ & 133.79 & 0.00 & 94.45 & 0.00 & 197.75 & 0.00 & 138.96 & 0.00 \\
        \midrule
        \multicolumn{9}{l}{\textnormal{\textit{GFlowNet, average}}} \\[-1pt]
        \textsc{Standalone} & 122.45 & 8.48 & 81.57 & 13.64 & 181.87 & 8.03 & 121.75 & 12.38 \\
        \textsc{N-FPT} \textnormal{(ours)} & 123.86 & 7.42 & 82.44 & 12.72 & 183.49 & 7.21 & 122.58 & 11.79 \\
        \textsc{NIC-FPT} \textnormal{(ours)} & \textbf{125.09} & \textbf{6.50} & \textbf{83.44} & \textbf{11.65} & \textbf{185.31} & \textbf{6.29} & \textbf{124.10} & \textbf{10.69} \\
        \addlinespace[2pt]
        \multicolumn{9}{l}{\textnormal{\textit{GFlowNet, best-of-20}}} \\[-1pt]
        \textsc{Standalone} & 125.06 & 6.52 & 84.43 & 10.61 & 184.86 & 6.52 & 125.56 & 9.64 \\
        \textsc{N-FPT} \textnormal{(ours)} & 126.30 & 5.60 & 85.14 & 9.86 & 186.30 & 5.79 & 126.28 & 9.12 \\
        \textsc{NRD-FPT} \textnormal{(ours)} & \textbf{128.32} & \textbf{4.09} & \textbf{86.31} & \textbf{8.61} & \textbf{189.26} & \textbf{4.29} & \textbf{127.63} & \textbf{8.16} \\
        \bottomrule
      \end{tabular}
    \end{adjustbox}
  }
\end{table}

\begin{table}[!ht]
  \centering
  \setlength{\tabcolsep}{0.0ex}
  \renewcommand{\arraystretch}{0.9}
  \caption{MIS on ER graphs with GFlowNet. Reporting conventions are in
  Appendix~\ref{app:inference-table-conventions}.}
  \label{tab:er}

  {\scshape\footnotesize
    \begin{adjustbox}{max width=\columnwidth}
      \begin{tabular}{@{}l*{4}{@{\hspace{1.5ex}}c@{\hspace{1.0ex}}c}@{}}
        \toprule
        \multirow{2}{*}{Method}
          & \multicolumn{2}{c}{SS} & \multicolumn{2}{c}{SD}
          & \multicolumn{2}{c}{LS} & \multicolumn{2}{c}{LD} \\
        \cmidrule(lr){2-3}\cmidrule(lr){4-5}\cmidrule(lr){6-7}\cmidrule(lr){8-9}
          & Size $\uparrow$ & Gap\% $\downarrow$ & Size $\uparrow$ & Gap\% $\downarrow$
          & Size $\uparrow$ & Gap\% $\downarrow$ & Size $\uparrow$ & Gap\% $\downarrow$ \\
        \midrule
        \textsc{Gurobi} \textnormal{(60s)} & 137.19 & 2.36 & 64.76 & 3.52 & 142.88 & 7.91 & 64.03 & 11.73 \\
        \textsc{Gurobi} \textnormal{(600s)}$^*$ & 140.51 & 0.00 & 67.12 & 0.00 & 155.16 & 0.00 & 72.54 & 0.00 \\
        \midrule
        \multicolumn{9}{l}{\textnormal{\textit{GFlowNet, average}}} \\[-1pt]
        \textsc{Standalone} & 133.96 & 4.66 & 63.42 & 5.51 & 150.62 & 2.93 & 71.19 & 1.86 \\
        \textsc{N-FPT} \textnormal{(ours)} & 134.17 & 4.51 & 63.47 & 5.44 & 150.72 & 2.86 & 71.20 & 1.85 \\
        \textsc{NIC-FPT} \textnormal{(ours)} & \textbf{136.95} & \textbf{2.53} & \textbf{64.22} & \textbf{4.32} & \textbf{154.25} & \textbf{0.59} & \textbf{71.80} & \textbf{1.02} \\
        \addlinespace[2pt]
        \multicolumn{9}{l}{\textnormal{\textit{GFlowNet, best-of-20}}} \\[-1pt]
        \textsc{Standalone} & 137.53 & 2.12 & 66.53 & 0.88 & 154.73 & 0.28 & \textbf{74.16} & \textbf{-2.23} \\
        \textsc{N-FPT} \textnormal{(ours)} & 137.74 & 1.97 & 66.57 & 0.82 & 154.80 & 0.23 & \textbf{74.16} & \textbf{-2.23} \\
        \textsc{NRD-FPT} \textnormal{(ours)} & \textbf{139.91} & \textbf{0.43} & \textbf{66.80} & \textbf{0.48} & \textbf{156.86} & \textbf{-1.10} & 73.85 & -1.81 \\
        \bottomrule
      \end{tabular}
    \end{adjustbox}
  }
\end{table}

On ER-LD (Table~\ref{tab:er}), \textsc{NRD-FPT} has a slightly lower
best-of-20 mean than \textsc{N-FPT}. Randomized deferral changes the neural
advice, so it need not improve every sampled result. Each returned solution
is nevertheless certified as best achievable under its own fixed advice;
the difference concerns advice quality, not the exactness of completion.
\begin{table}[!ht]
  \centering
  \setlength{\tabcolsep}{0.0ex}
  \renewcommand{\arraystretch}{0.9}
  \caption{MIS on HK graphs with GFlowNet. Reporting conventions are in
  Appendix~\ref{app:inference-table-conventions}.}
  \label{tab:hk}

  {\scshape\footnotesize
    \begin{adjustbox}{max width=\columnwidth}
      \begin{tabular}{@{}l*{4}{@{\hspace{1.5ex}}c@{\hspace{1.0ex}}c}@{}}
        \toprule
        \multirow{2}{*}{Method}
          & \multicolumn{2}{c}{SS} & \multicolumn{2}{c}{SD}
          & \multicolumn{2}{c}{LS} & \multicolumn{2}{c}{LD} \\
        \cmidrule(lr){2-3}\cmidrule(lr){4-5}\cmidrule(lr){6-7}\cmidrule(lr){8-9}
          & Size $\uparrow$ & Gap\% $\downarrow$ & Size $\uparrow$ & Gap\% $\downarrow$
          & Size $\uparrow$ & Gap\% $\downarrow$ & Size $\uparrow$ & Gap\% $\downarrow$ \\
        \midrule
        \textsc{Gurobi} \textnormal{(60s)} & 332.69 & 0.00 & 225.52 & 0.80 & 488.54 & 0.01 & 327.94 & 1.36 \\
        \textsc{Gurobi} \textnormal{(600s)}$^*$ & 332.70 & 0.00 & 227.33 & 0.00 & 488.57 & 0.00 & 332.47 & 0.00 \\
        \midrule
        \multicolumn{9}{l}{\textnormal{\textit{GFlowNet, average}}} \\[-1pt]
        \textsc{Standalone} & 331.74 & 0.29 & 219.89 & 3.27 & 486.20 & 0.49 & 323.81 & 2.61 \\
        \textsc{N-FPT} \textnormal{(ours)} & 331.83 & 0.26 & 221.30 & 2.65 & 486.80 & 0.36 & 325.14 & 2.20 \\
        \textsc{NIC-FPT} \textnormal{(ours)} & \textbf{332.05} & \textbf{0.19} & \textbf{222.96} & \textbf{1.92} & \textbf{487.33} & \textbf{0.25} & \textbf{327.24} & \textbf{1.57} \\
        \addlinespace[2pt]
        \multicolumn{9}{l}{\textnormal{\textit{GFlowNet, best-of-20}}} \\[-1pt]
        \textsc{Standalone} & 332.10 & 0.18 & 221.98 & 2.35 & 486.97 & 0.33 & 326.33 & 1.85 \\
        \textsc{N-FPT} \textnormal{(ours)} & 332.16 & 0.16 & 223.32 & 1.76 & 487.45 & 0.23 & 327.53 & 1.49 \\
        \textsc{NRD-FPT} \textnormal{(ours)} & \textbf{332.57} & \textbf{0.04} & \textbf{226.37} & \textbf{0.42} & \textbf{488.30} & \textbf{0.05} & \textbf{331.23} & \textbf{0.38} \\
        \bottomrule
      \end{tabular}
    \end{adjustbox}
  }
\end{table}

\begin{table}[!ht]
  \centering
  \setlength{\tabcolsep}{0.0ex}
  \renewcommand{\arraystretch}{0.9}
  \caption{MIS on BA graphs with GFlowNet. Reporting conventions are in
  Appendix~\ref{app:inference-table-conventions}.}
  \label{tab:ba}

  {\scshape\footnotesize
    \begin{adjustbox}{max width=\columnwidth}
      \begin{tabular}{@{}l*{4}{@{\hspace{1.5ex}}c@{\hspace{1.0ex}}c}@{}}
        \toprule
        \multirow{2}{*}{Method}
          & \multicolumn{2}{c}{SS} & \multicolumn{2}{c}{SD}
          & \multicolumn{2}{c}{LS} & \multicolumn{2}{c}{LD} \\
        \cmidrule(lr){2-3}\cmidrule(lr){4-5}\cmidrule(lr){6-7}\cmidrule(lr){8-9}
          & Size $\uparrow$ & Gap\% $\downarrow$ & Size $\uparrow$ & Gap\% $\downarrow$
          & Size $\uparrow$ & Gap\% $\downarrow$ & Size $\uparrow$ & Gap\% $\downarrow$ \\
        \midrule
        \textsc{Gurobi} \textnormal{(60s)} & 408.10 & 0.00 & 182.61 & 1.66 & 601.55 & 0.00 & 264.69 & 2.94 \\
        \textsc{Gurobi} \textnormal{(600s)}$^*$ & 408.10 & 0.00 & 185.69 & 0.00 & 601.55 & 0.00 & 272.70 & 0.00 \\
        \midrule
        \multicolumn{9}{l}{\textnormal{\textit{GFlowNet, average}}} \\[-1pt]
        \textsc{Standalone} & 408.02 & 0.02 & 177.18 & 4.58 & 601.50 & \textbf{0.01} & 261.50 & 4.11 \\
        \textsc{N-FPT} \textnormal{(ours)} & 408.02 & 0.02 & 178.20 & 4.03 & 601.50 & \textbf{0.01} & 262.92 & 3.59 \\
        \textsc{NIC-FPT} \textnormal{(ours)} & \textbf{408.04} & \textbf{0.01} & \textbf{180.95} & \textbf{2.55} & \textbf{601.51} & \textbf{0.01} & \textbf{266.62} & \textbf{2.23} \\
        \addlinespace[2pt]
        \multicolumn{9}{l}{\textnormal{\textit{GFlowNet, best-of-20}}} \\[-1pt]
        \textsc{Standalone} & 408.05 & 0.01 & 180.09 & 3.01 & 601.51 & 0.01 & 265.17 & 2.76 \\
        \textsc{N-FPT} \textnormal{(ours)} & 408.05 & 0.01 & 181.01 & 2.52 & 601.52 & \textbf{0.00} & 266.44 & 2.30 \\
        \textsc{NRD-FPT} \textnormal{(ours)} & \textbf{408.09} & \textbf{0.00} & \textbf{184.15} & \textbf{0.83} & \textbf{601.54} & \textbf{0.00} & \textbf{270.28} & \textbf{0.89} \\
        \bottomrule
      \end{tabular}
    \end{adjustbox}
  }
\end{table}

\begin{table}[!ht]
  \centering
  \setlength{\tabcolsep}{0.0ex}
  \renewcommand{\arraystretch}{0.9}
  \caption{MVC on Reg graphs with GFlowNet. Reporting conventions are in
  Appendix~\ref{app:inference-table-conventions}.}
  \label{tab:mvc-reg}

  {\scshape\footnotesize
    \begin{adjustbox}{max width=\columnwidth}
      \begin{tabular}{@{}l*{4}{@{\hspace{1.5ex}}c@{\hspace{1.0ex}}c}@{}}
        \toprule
        \multirow{2}{*}{Method}
          & \multicolumn{2}{c}{SS} & \multicolumn{2}{c}{SD}
          & \multicolumn{2}{c}{LS} & \multicolumn{2}{c}{LD} \\
        \cmidrule(lr){2-3}\cmidrule(lr){4-5}\cmidrule(lr){6-7}\cmidrule(lr){8-9}
          & Size $\downarrow$ & Gap\% $\downarrow$ & Size $\downarrow$ & Gap\% $\downarrow$
          & Size $\downarrow$ & Gap\% $\downarrow$ & Size $\downarrow$ & Gap\% $\downarrow$ \\
        \midrule
        \textsc{Gurobi} \textnormal{(60s)} & 566.33 & 1.47 & 678.80 & 0.50 & 743.05 & 2.35 & 885.11 & 0.94 \\
        \textsc{Gurobi} \textnormal{(600s)}$^*$ & 558.10 & 0.00 & 675.45 & 0.00 & 725.96 & 0.00 & 876.85 & 0.00 \\
        \midrule
        \multicolumn{9}{l}{\textnormal{\textit{GFlowNet, average}}} \\[-1pt]
        \textsc{Standalone} & 561.59 & 0.62 & 680.14 & 0.70 & 729.77 & 0.52 & 881.69 & \textbf{0.55} \\
        \textsc{N-FPT} \textnormal{(ours)} & 561.50 & 0.61 & 680.13 & 0.69 & 729.67 & 0.51 & \textbf{881.68} & \textbf{0.55} \\
        \textsc{NIC-FPT} \textnormal{(ours)} & \textbf{561.19} & \textbf{0.55} & \textbf{680.07} & \textbf{0.68} & \textbf{729.44} & \textbf{0.48} & 881.69 & \textbf{0.55} \\
        \addlinespace[2pt]
        \multicolumn{9}{l}{\textnormal{\textit{GFlowNet, best-of-20}}} \\[-1pt]
        \textsc{Standalone} & 557.50 & -0.11 & 675.87 & 0.06 & 725.10 & -0.12 & 876.80 & -0.01 \\
        \textsc{N-FPT} \textnormal{(ours)} & 557.42 & -0.12 & 675.86 & 0.06 & 724.97 & -0.14 & 876.78 & -0.01 \\
        \textsc{NRD-FPT} \textnormal{(ours)} & \textbf{554.15} & \textbf{-0.71} & \textbf{673.58} & \textbf{-0.28} & \textbf{720.54} & \textbf{-0.75} & \textbf{874.35} & \textbf{-0.29} \\
        \bottomrule
      \end{tabular}
    \end{adjustbox}
  }
\end{table}

\begin{table}[!ht]
  \centering
  \setlength{\tabcolsep}{0.0ex}
  \renewcommand{\arraystretch}{0.9}
  \caption{MaxCut on Reg graphs with GFlowNet. Reporting conventions are in
  Appendix~\ref{app:inference-table-conventions}.}
  \label{tab:mcut-reg}

  {\scshape\footnotesize
    \begin{adjustbox}{max width=\columnwidth}
      \begin{tabular}{@{}l*{4}{@{\hspace{1.5ex}}c@{\hspace{1.0ex}}c}@{}}
        \toprule
        \multirow{2}{*}{Method}
          & \multicolumn{2}{c}{SS} & \multicolumn{2}{c}{SD}
          & \multicolumn{2}{c}{LS} & \multicolumn{2}{c}{LD} \\
        \cmidrule(lr){2-3}\cmidrule(lr){4-5}\cmidrule(lr){6-7}\cmidrule(lr){8-9}
          & Size $\uparrow$ & Gap\% $\downarrow$ & Size $\uparrow$ & Gap\% $\downarrow$
          & Size $\uparrow$ & Gap\% $\downarrow$ & Size $\uparrow$ & Gap\% $\downarrow$ \\
        \midrule
        \textsc{Gurobi} \textnormal{(60s)} & 2006.10 & 0.00 & 4644.71 & 0.00 & 2601.79 & 0.00 & 6006.83 & 0.00 \\
        \textsc{Gurobi} \textnormal{(600s)}$^*$ & 2006.15 & 0.00 & 4644.71 & 0.00 & 2601.82 & 0.00 & 6006.83 & 0.00 \\
        \midrule
        \multicolumn{9}{l}{\textnormal{\textit{GFlowNet, average}}} \\[-1pt]
        \textsc{Standalone} & 1944.50 & 3.07 & 4458.06 & 4.02 & 2582.00 & 0.76 & 5767.26 & 3.99 \\
        \textsc{N-FPT} \textnormal{(ours)} & 1967.72 & 1.92 & 4474.58 & 3.66 & 2587.64 & 0.55 & 5792.78 & 3.56 \\
        \textsc{NIC-FPT} \textnormal{(ours)} & \textbf{1996.30} & \textbf{0.49} & \textbf{4490.85} & \textbf{3.31} & \textbf{2608.50} & \textbf{-0.26} & \textbf{5824.44} & \textbf{3.04} \\
        \addlinespace[2pt]
        \multicolumn{9}{l}{\textnormal{\textit{GFlowNet, best-of-20}}} \\[-1pt]
        \textsc{Standalone} & 1963.73 & 2.11 & 4492.65 & 3.27 & 2601.64 & 0.01 & 5804.41 & 3.37 \\
        \textsc{N-FPT} \textnormal{(ours)} & 1986.51 & 0.98 & 4509.91 & 2.90 & 2606.80 & -0.19 & 5830.97 & 2.93 \\
        \textsc{NRD-FPT} \textnormal{(ours)} & \textbf{2007.57} & \textbf{-0.07} & \textbf{4544.66} & \textbf{2.15} & \textbf{2618.95} & \textbf{-0.66} & \textbf{5894.62} & \textbf{1.87} \\
        \bottomrule
      \end{tabular}
    \end{adjustbox}
  }
\end{table}

\FloatBarrier
\subsection{More Results on Generalization}
\label{app:generalization}
\begin{figure}[!htbp]
  \centering
  \includegraphics[width=0.745\linewidth]{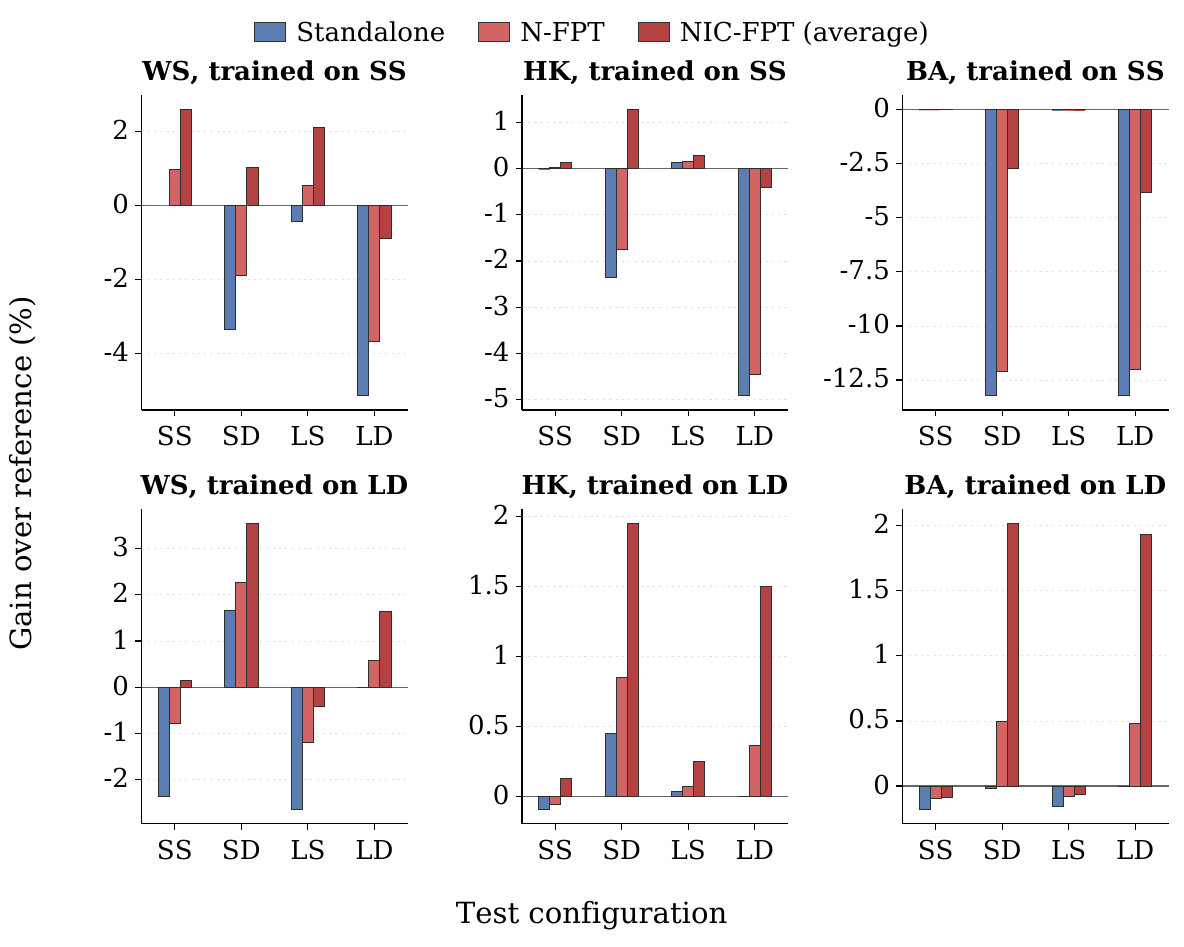}
  \caption{Within-family generalization: percentage gains
  in average solution size. Higher is better.}
  \label{fig:in-class-gen}
\end{figure}

\begin{figure}[!htbp]
  \centering
  \includegraphics[width=0.745\linewidth]{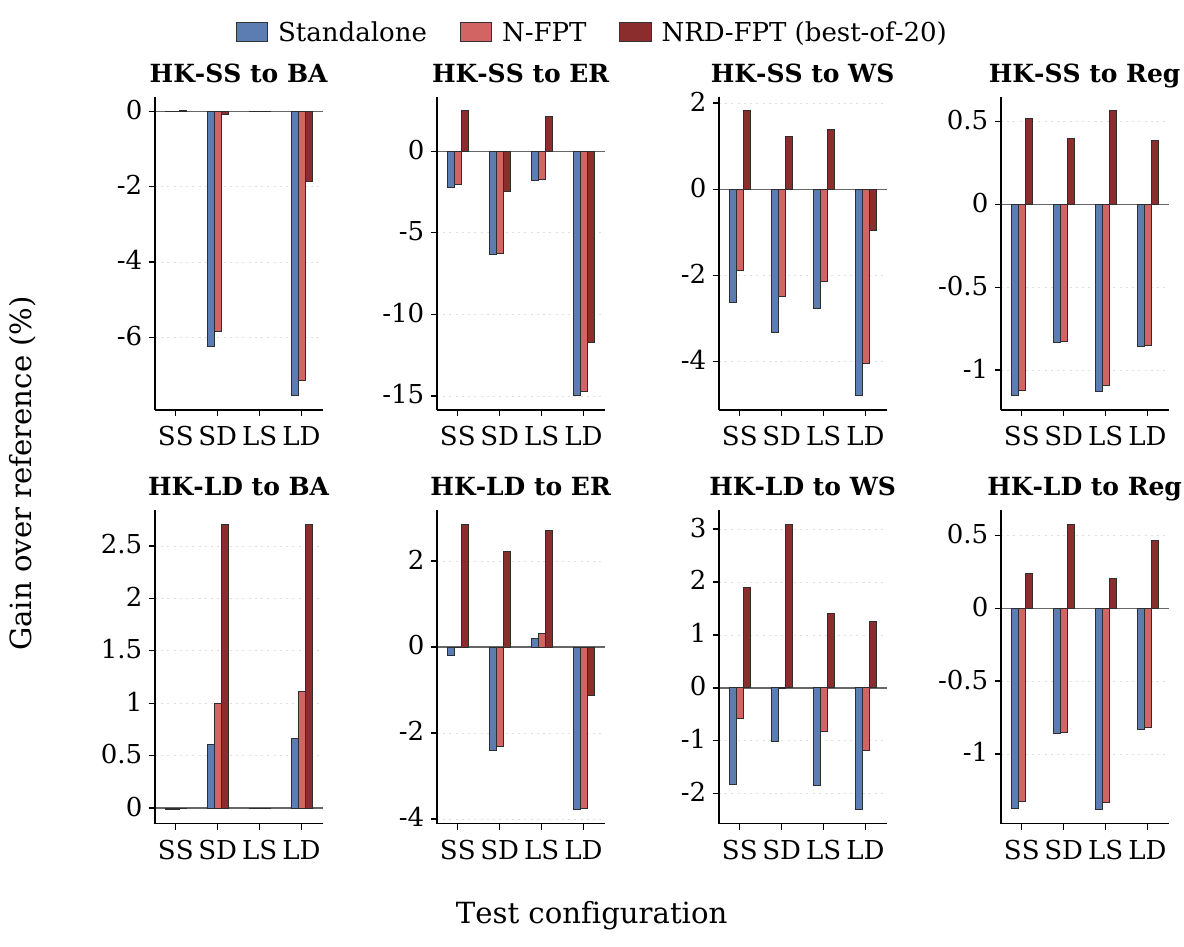}
  \caption{Cross-family generalization from HK graphs: percentage gains
  in best-of-20 solution size. Higher is better.}
  \label{fig:cross-hk}
\end{figure}

\begin{figure}[!htbp]
  \centering
  \includegraphics[width=0.745\linewidth]{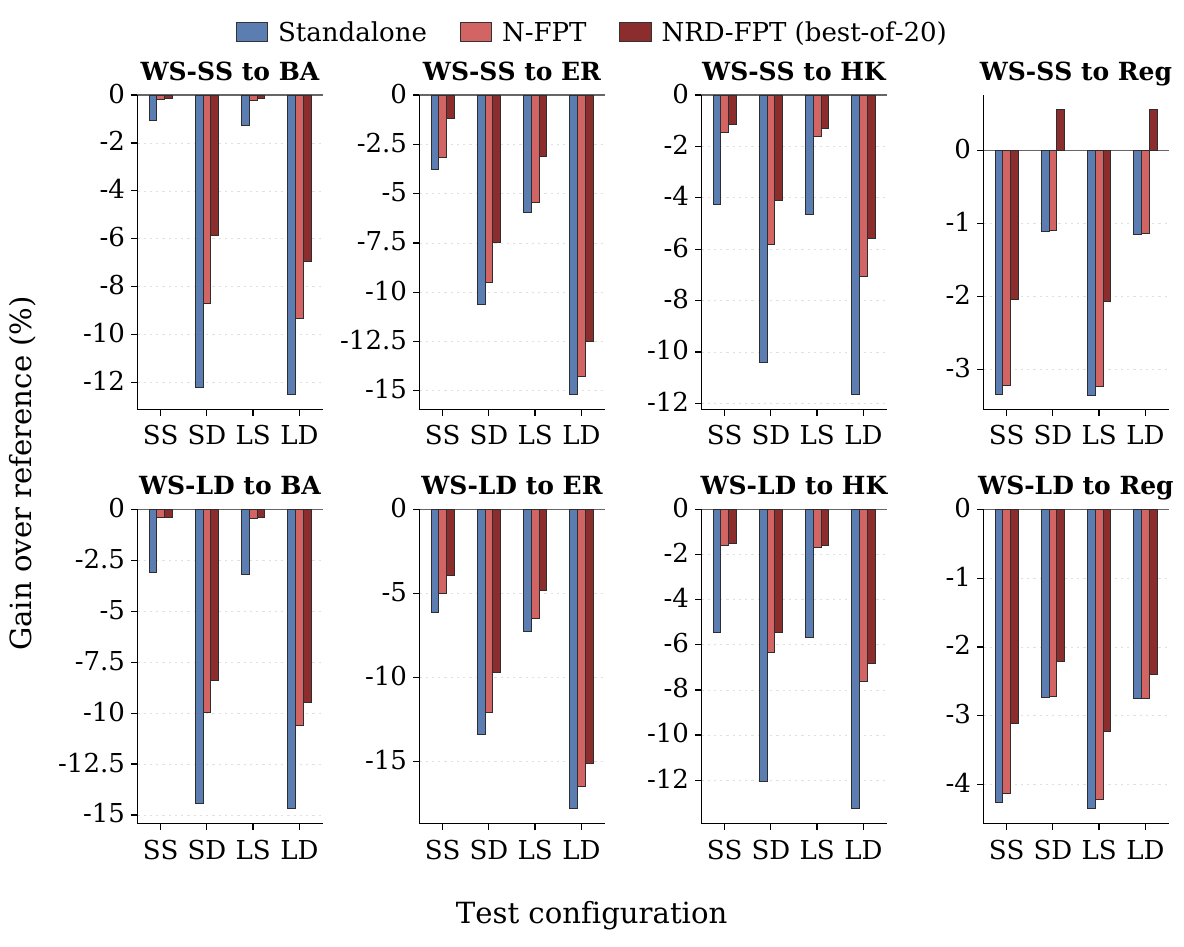}
  \caption{Cross-family generalization from WS graphs: percentage gains
  in best-of-20 solution size. Higher is better.}
  \label{fig:cross-ws}
\end{figure}

\begin{figure}[!htbp]
  \centering
  \includegraphics[width=0.745\linewidth]{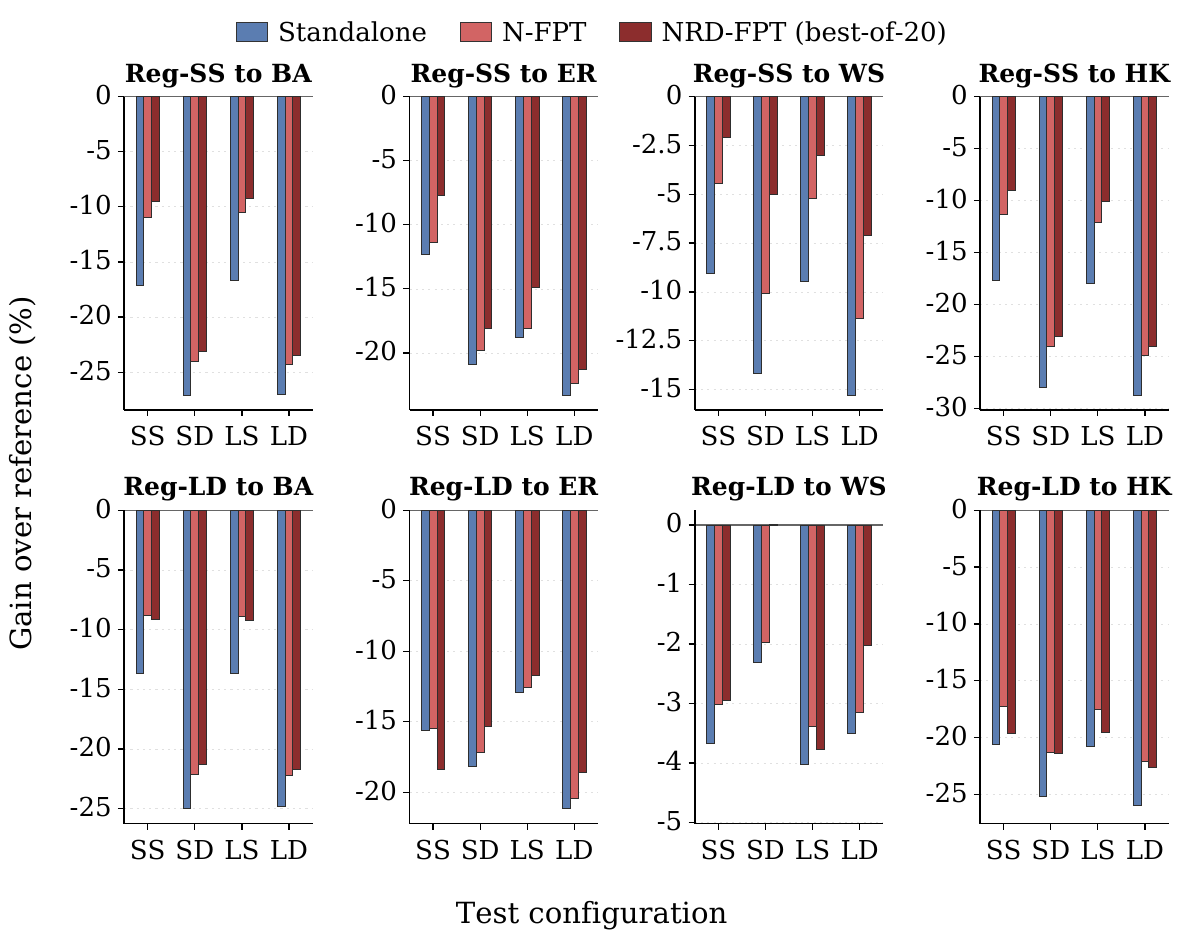}
  \caption{Cross-family generalization from Reg graphs: percentage gains
  in best-of-20 solution size. Higher is better.}
  \label{fig:cross-reg}
\end{figure}

Figures~\ref{fig:in-class-gen}--\ref{fig:cross-reg} extend
Figure~\ref{fig:generalization} with LD-trained models and further source
families. The reference and sampling
measures follow Appendix~\ref{app:generalization-protocol}. Each figure
places SS-trained models above LD-trained models.

All plots report the gain over the target-trained neural reference,
$100(B/A-1)$, as in Figure~\ref{fig:generalization}; higher is better.
Within-family results use average solution size and compare
\textsc{Standalone}, \textsc{N-FPT}, and \textsc{NIC-FPT}.
Cross-family results use best-of-20 and compare \textsc{Standalone},
\textsc{N-FPT}, and \textsc{NRD-FPT}. These gains are not certified
optimality gaps.

The additional source families show that the size of the transfer loss
depends on where the neural model was trained. Exact completion can improve
transferred solutions across these families. The variants do not remove every deficit relative to the
target-trained reference: the quality of the fixed advice still matters.
SS and LD bars also have different target-trained references, so their
heights compare relative performance, not absolute solution sizes.

\FloatBarrier
\subsection{Further Results on Learned Completion}
\label{app:nc-completion-quality}
For each decoded solution $S$, we compute the percentage gap
$100(1-f_G(S)/V_G(a_0))$, then average over test graphs and training seeds.
For the paired comparisons, we average each method's solution size over its
four trained models on each test graph, then record a win, tie, or loss for
RL with \textsc{N-FPT} feedback relative to RL. The plotted $p$-values are
from two-sided exact sign tests on non-tied graphs, testing whether either
method is equally likely to win.

\begin{figure}[!ht]
  \centering
  \includegraphics[width=0.846\linewidth]{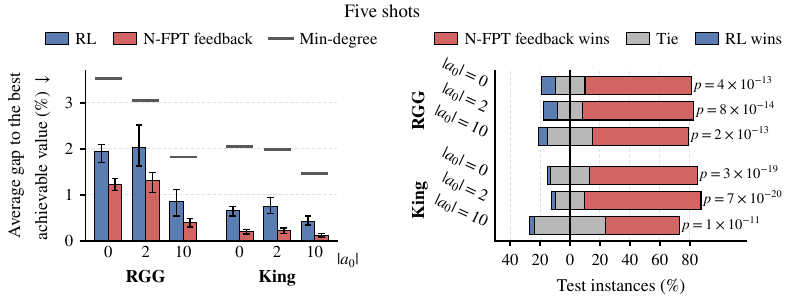}
  \caption{Five-shot completion quality; visual conventions follow
  Figure~\ref{fig:nc-completion-quality}.}
  \label{fig:nc-completion-quality-five-shot}
\end{figure}

Under one-shot decoding (Figure~\ref{fig:nc-completion-quality}), feedback
improves 57--75 of the 100 test instances per setting and worsens only 6--19.
Figure~\ref{fig:nc-completion-quality-five-shot} reports the corresponding
five-shot results: each trained model returns its best of five sampled
completions, with all other settings unchanged. Feedback improves 49--78
instances per setting and worsens only 2--9. Under both decoding budgets,
feedback lowers the mean gap in every setting, and all sign tests give
$p<10^{-7}$. The advantage of feedback therefore persists when both methods
can sample several completions.

\noindent\textbf{Preserving Optimal Completion during Training.}
Figure~\ref{fig:nc-feedback-training} tracks how often sampled trajectories
lose optimal completion and where the first harmful selection occurs.
Early in training, nearly all sampled trajectories miss the best achievable
value, yet they already provide exact corrective lessons. As training
progresses, fewer trajectories lose optimal completion and, among those that
still do, the first harmful selection generally occurs later. These trends
support the decision-level benefit observed on test trajectories in
Figure~\ref{fig:nc-optimum-survival}.

\FloatBarrier

\section{Reproduction Notes}
\subsection{Hardware}
The original experiments used an Intel Xeon Platinum 8360Y CPU at 2.40\,GHz,
up to 32\,GB of RAM, and a single NVIDIA A100-SXM4 GPU. Additional
experiments ran on NCI's Gadi supercomputer, primarily using CPUs and
NVIDIA V100 GPUs, consuming approximately 20,000 service units. We also
used a Slurm-managed cluster, consuming approximately 20,000 service units
under that system's accounting.

\subsection{Inference Datasets} \label{appendix:dataset}
We vary graph families and cross graph size with density within each family
to test whether gains depend on a particular structure or distribution.
The families include Erd\H{o}s--R\'enyi (ER)~\citep{erdds1959random},
Barab\'asi--Albert (BA)~\citep{barabasi1999emergence},
Watts--Strogatz (WS)~\citep{watts1998collective}, random regular
(Reg)~\citep{steger1999generating}, and Holme--Kim (HK) graphs.
We generate them with NetworkX~\citep{networkx}, following the parameters
in Table~\ref{tab:dataset}. Each configuration contains 100 training and
100 test graphs generated with distinct random seeds. Vertex counts are
sampled uniformly from the listed ranges; the other parameter names follow
NetworkX. We remove self-loops and isolated vertices and relabel vertices
with consecutive integers for implementation purposes.

The four configurations are small sparse (SS), small dense (SD), large sparse
(LS), and large dense (LD). For the Reg graphs in Tables~\ref{tab:reg}
and~\ref{tab:cross-problem}, small
instances have 800--900 vertices and large instances have 1,000--1,200;
the sparse and dense settings use degrees 6 and 16, respectively.
Cross-size tests hold the other generation parameters fixed;
cross-distribution tests change those parameters while retaining the same
vertex-count range.
Vertex count alone does not determine the difficulty of exact completion:
the residual width controls its exponential cost, while the modulator
determines how many decisions rely on neural advice.
\begin{table}[!ht]
  \centering
  \setlength{\tabcolsep}{0.0ex}
  \renewcommand{\arraystretch}{0.9}
  \caption{Graph-generation settings and mean vertex/edge counts for the
  original datasets. Parameter names follow NetworkX~\citep{networkx} and
  are independent of the notation elsewhere in the paper.}
  \label{tab:dataset}

  {\scshape\footnotesize
    \begin{adjustbox}{max width=\columnwidth}
      \begin{tabular}{@{}l@{\hspace{3ex}}c@{\hspace{3ex}}c@{}}
        \toprule
        Dataset & Configuration & Mean $(|V|,|E|)$ \\
        \midrule
        $\mathrm{ER}_{\mathrm{SS}}$ & $\left[700,800\right], p=0.03$ & $(748.05,8408.72)$ \\
        $\mathrm{ER}_{\mathrm{SD}}$ & $\left[700,800\right], p=0.08$ & $(750.54,22551.89)$ \\
        $\mathrm{ER}_{\mathrm{LS}}$ & $\left[1000,1200\right], p=0.03$ & $(1102.32,18258.46)$ \\
        $\mathrm{ER}_{\mathrm{LD}}$ & $\left[1000,1200\right], p=0.08$ & $(1103.03,48750.79)$ \\
        \addlinespace[2pt]
        $\mathrm{BA}_{\mathrm{SS}}$ & $\left[700,800\right], m=3$ & $(748.46,1840.49)$ \\
        $\mathrm{BA}_{\mathrm{SD}}$ & $\left[700,800\right], m=15$ & $(749.43,10655.76)$ \\
        $\mathrm{BA}_{\mathrm{LS}}$ & $\left[1000,1200\right], m=3$ & $(1101.31,2702.49)$ \\
        $\mathrm{BA}_{\mathrm{LD}}$ & $\left[1000,1200\right], m=15$ & $(1090.02,15538.75)$ \\
        \addlinespace[2pt]
        $\mathrm{WS}_{\mathrm{SS}}$ & $\left[700,800\right], k=15, p=0.1$ & $(748.05,5236.35)$ \\
        $\mathrm{WS}_{\mathrm{SD}}$ & $\left[700,800\right], k=25, p=0.1$ & $(750.54,9006.48)$ \\
        $\mathrm{WS}_{\mathrm{LS}}$ & $\left[1000,1200\right], k=15, p=0.1$ & $(1102.32,7716.24)$ \\
        $\mathrm{WS}_{\mathrm{LD}}$ & $\left[1000,1200\right], k=25, p=0.1$ & $(1103.03,13236.36)$ \\
        \addlinespace[2pt]
        $\mathrm{Reg}_{\mathrm{SS}}$ & $\left[800,900\right], d=6$ & $(848.05,2544.15)$ \\
        $\mathrm{Reg}_{\mathrm{SD}}$ & $\left[800,900\right], d=16$ & $(850.54,6804.32)$ \\
        $\mathrm{Reg}_{\mathrm{LS}}$ & $\left[1000,1200\right], d=6$ & $(1102.32,3306.96)$ \\
        $\mathrm{Reg}_{\mathrm{LD}}$ & $\left[1000,1200\right], d=16$ & $(1103.03,8824.24)$ \\
        \addlinespace[2pt]
        $\mathrm{HK}_{\mathrm{SS}}$ & $\left[700,800\right], m\in\left[3,7\right],p=0.3$ & $(748.46,3299.16)$ \\
        $\mathrm{HK}_{\mathrm{SD}}$ & $\left[700,800\right], m\in\left[10,15\right],p=0.3$ & $(746.66,8743.89)$ \\
        $\mathrm{HK}_{\mathrm{LS}}$ & $\left[1000,1200\right], m\in\left[3,7\right],p=0.3$ & $(1101.31,4941.53)$ \\
        $\mathrm{HK}_{\mathrm{LD}}$ & $\left[1000,1200\right], m\in\left[10,15\right],p=0.3$ & $(1096.27,13172.38)$ \\
        \bottomrule
      \end{tabular}
    \end{adjustbox}
  }
\end{table}

\subsection{Inference Evaluation Protocol}
\label{app:inference-evaluation}
Tables~\ref{tab:reg} and~\ref{tab:cross-problem} use 100 test graphs per
configuration and three training seeds for each neural model. We retain
20 final samples per graph. The average result is their mean objective
value; best-of-20 takes the largest value for MIS and MaxCut and the
smallest for MVC. Both are then averaged over test graphs and training
seeds. MaxCut values count crossing edges; MIS and MVC values count
selected vertices.
\textsc{Standalone} denotes the trained neural solver alone. Each
\textsc{N-FPT} result exactly completes the modulator advice from its paired
neural sample. Residual-width settings and the exact construction are in
Appendix~\ref{sec:td-tmgtd-tdpa}.
\textsc{NIC-FPT} and \textsc{NRD-FPT} use the five-round consolidated-prediction
protocols in Appendix~\ref{sec:icl-details}. They use additional
inference sampling, not additional training. Training details are in
Appendices~\ref{sec:gflownet-training} and~\ref{sec:gnnrl-training}.

For the MIS comparison in Table~\ref{tab:reg}, GFlowNet and Gurobi use the
original graph instances; GNN+RL and KaMIS use newly generated instances
with the same configuration ranges. Within each
neural model, the standalone solver and all \textsc{N-FPT} variants share
checkpoints and test graphs. Thus, the paired evidence for improvement is within each
neural model, not a ranking of solvers across independently generated datasets.

Gurobi provides a reference across the optimization
problems~\citep{mittelmann2020benchmarking}. We use 60s and 600s limits with
a single thread, retaining the best incumbent when a run reaches its limit,
and average over 10 runs with different random seeds. KaMIS runs for
600 seconds per instance on the GNN+RL MIS test graphs.

Table~\ref{tab:reg} reports the reference gap as $100(1-x/b)$, where $x$ is the reported
mean size and $b$ is the original Gurobi (600s) reference mean marked with
$*$. These values are not certified optima, nor are they paired references
for the newly generated GNN+RL instances. Negative gaps indicate a size
above the reference mean, not improvement over a proven optimum. Arrows
indicate the preferred direction: larger size and smaller reference gap.
Table~\ref{tab:cross-problem} reports objective values only; the full
tables and problem-specific gap definitions are in
Appendix~\ref{app:inference-table-conventions}.

\paragraph{Generalization.}
\label{app:generalization-protocol}
The generalization experiments use GFlowNet on MIS. Within each of the WS,
HK, and BA families, we train on SS or LD and test on all four configurations.
The matching training configuration is an in-distribution control; the
others change graph size, density, or both. Across families, we train on HK,
WS, or Reg graphs, using SS or LD, and test on all four configurations of
each other family. Model weights remain fixed throughout evaluation.

For each target family and size--density configuration, the reference $A$
is the result of a standalone GFlowNet trained specifically on that
configuration. The compared value $B$ comes from the source-trained model,
either standalone or within an \textsc{N-FPT} method. Within-family results
use average solution size; across-family results use best-of-20, with the
same measure for $A$ and $B$. Figure~\ref{fig:generalization} reports
$100(B/A-1)$: zero matches the target-trained reference, and positive values
indicate larger solutions. This is a comparison with a neural reference,
not an optimality gap. The figure uses the SS-trained models, with HK as the
cross-family source; Appendix~\ref{app:generalization} includes LD-trained
models and the other source families.

\subsection{Problem Definitions and Gurobi} \label{appendix:more-gurobi}
For MIS, the valid solution family and objective are
\[
  \mathcal{F}_{\mathrm{MIS}}(G)
  =\{S\subseteq V:\forall (u,v)\in E,\ u\notin S\ \text{or}\ v\notin S\},
  \qquad \max_{S\in\mathcal{F}_{\mathrm{MIS}}(G)}|S|.
\]
For MVC, they are
\[
  \mathcal{F}_{\mathrm{MVC}}(G)
  =\{S\subseteq V:\forall (u,v)\in E,\ u\in S\ \text{or}\ v\in S\},
  \qquad \min_{S\in\mathcal{F}_{\mathrm{MVC}}(G)}|S|.
\]
For MaxCut, every subset $S\subseteq V$ defines the cut
$\delta(S)=\{(u,v)\in E:|\{u,v\}\cap S|=1\}$, and the objective is
$\max_{S\subseteq V}|\delta(S)|$.

\subsection{More Details on GFlowNet Training.} \label{sec:gflownet-training}
We train GFlowNet following prior work~\citep{zhang2023letflowstellsolving}. In addition, we make two modifications to improve performance. First, we apply the non-annealing setting during training. Second, we replace the average pooling layer from the original policy network
with {\it a sum pooling layer}. The previous study limits training to 20 epochs, with convergence often occurring within 5. However, due to the more dynamic and heterogeneous nature of our dataset, and to further investigate the best GFlowNet can achieve, we increase the training budget to 50 epochs for each configuration. Models are saved at every epoch, and evaluation is conducted using the model that achieves the best performance. We train each configuration with three random seeds and report the mean performance across these runs.

\subsection{GNN+RL Training and Evaluation}
\label{sec:gnnrl-training}
The GNN+RL policy is a five-layer graph attention network with hidden
dimension 256 and four attention heads. It observes each vertex's decision
status and the number of undecided neighbors. At each step, it samples an
undecided vertex, selects it, and excludes its neighbors. We train with
REINFORCE on 100 training graphs per configuration for 50 epochs, using
batches of up to 64 graphs and four sampled solutions per graph. Each
solution's advantage is its size minus the mean size of the other three
samples from the same graph. We use Adam with learning rate $10^{-3}$ and
gradient clipping at norm 1. Results use the final checkpoint from each of
three training seeds and 20 sampled solutions per test graph.

\subsection{Neural Completion Training and Evaluation}
\label{app:nc-training-evaluation}
The learning experiments in Section~\ref{sec:learning} use MIS on random
geometric graphs (RGG) and king's graphs with random vertex deletions (King),
each with about 300 vertices. Each family has 500 training graphs,
50 validation graphs, and 100 test graphs. We draw a target size
$n_{\mathrm{target}}$ uniformly from the integers 270--330. RGG places this
many points uniformly in the unit square and joins points within distance
0.103. King starts from a square grid of side length
$\operatorname{round}(\sqrt{n_{\mathrm{target}}/0.7})$, retains each vertex
independently with probability 0.7, and joins retained vertices that are
horizontal, vertical, or diagonal neighbors. We remove isolated vertices
and self-loops and relabel vertices consecutively.

The three starting-state settings use the same graphs and contain zero,
two, or ten initial selections. For each graph and setting, we generate
one starting assignment by uniformly sampling an undecided vertex at each
selection and excluding its neighbors. This assignment stays fixed across
epochs and is shared by both methods.

Both RL and RL with \textsc{N-FPT} feedback use a five-layer graph attention
network with hidden dimension 256 and four attention heads of dimension 64.
The inputs are vertex-status embeddings and projected log-transformed
counts of undecided neighbors. The network uses residual connections and
ELU activations, without dropout or normalization layers, followed by a
two-layer output network with a ReLU activation.
Policies are trained from scratch for 20 epochs with matched seeds 1--4.
We train separate policies for each graph family and starting-state setting.

We use Adam with learning rate $10^{-4}$ and gradient-norm clipping at 1.
Each epoch visits all 500 training assignments in random order, with four
sampled trajectories and one optimizer update per assignment, giving
10,000 updates per run. Gradient computation processes at most 64 graph-state
copies per network pass. The feedback weight in
Equation~\ref{eq:nfpt-feedback-loss} is $\lambda=1$; the RL baseline omits
that term. Architecture, training graphs, starting assignments, and
optimizer-update counts are otherwise matched. After each epoch, we evaluate
greedy completion on all 50 validation graphs and retain the checkpoint
with the largest mean solution size for testing.

For these runs, exact feedback is prepared at the given starting assignments
when affordable, corresponding to $s=0$ in Section~\ref{sec:learning}.
We cap the exact computation at 200,000 valid assignments per bag and the
stored optimal-completion representation at 2,000,000 bag states per
starting assignment. The representations are cached and reused across
epochs. Unavailable feedback is omitted; these graphs remain in RL
training rather than being removed from the dataset.

Gurobi proves the reference value $V_G(a_0)$ for every test graph under
the same fixed assignment. These references only assess the outputs; each
trained model constructs its test completions without an exact solver.
One-shot decoding selects the highest-probability vertex at each step;
five-shot decoding returns the best of five sampled
completions. Every setting is evaluated against its own best achievable
value. We report optimal-completion frequency, mean percentage gap, and
paired instance-level comparisons; aggregation and statistical tests are
specified in Appendix~\ref{app:nc-completion-quality}.

\paragraph{Trajectory diagnostics.}
Figure~\ref{fig:nc-feedback-training} uses the trajectories sampled during
each training epoch for which exact feedback is available at $a_0$
($s=0$). For each suboptimal trajectory, we divide the index
of its first harmful selection by its number of additional selections and
report the mean as a percentage. The failure rate is the fraction of all
sampled trajectories from these starts that end below $V_G(a_0)$.
Starting assignments without
the stored representation are excluded from both diagnostics.

Figure~\ref{fig:nc-optimum-survival} replays the one-shot test trajectories
through the same type of stored representation. Completion progress is the
fraction of a trajectory's additional selections, not the fraction of graph
vertices decided. At each progress level, we count the test graphs on which
no harmful selection has yet occurred. Each curve therefore starts at
100\% and ends at the corresponding one-shot optimal-completion frequency.
All replayed solution sizes match the original evaluation. For these test
diagnostics, exact computation only diagnoses completed trajectories; it
neither chooses nor changes their selections. In both figures, lines average the four training seeds
and shaded bands span their minimum and maximum.

\subsection{Tree Decomposition, Modulator, and Exact Completion}
\label{sec:td-tmgtd-tdpa}
We construct the initial tree decomposition using the min-degree heuristic.
For a vertex $u$, let $N(u)$ and $N[u]=N(u)\cup\{u\}$ denote its open and
closed neighborhoods. As shown in Algorithm~\ref{alg:min-degree}, the heuristic
repeatedly selects a minimum-degree vertex $u$, adds fill edges that make
$N(u)$ a clique, creates the bag $N[u]$, and removes $u$. This produces a
valid decomposition efficiently, although not necessarily one of minimum
width~\citep{Bannach2017JdrasilAM}.
\begin{algorithm}[!ht]
  \caption{Min-degree tree-decomposition heuristic}
  \label{alg:min-degree}
  \begin{algorithmic}[1]
    \Require Graph $G=(V,E)$
    \Ensure A valid tree decomposition $\mathcal{T}=(T,\{X_t\}_{t\in V(T)})$
    \State $H\gets G$; initialize an empty elimination sequence $L$
    \While{$V(H)\neq\emptyset$}
      \State Choose a minimum-degree vertex $u\in V(H)$
      \State Record the bag $X_u=\{u\}\cup N_H(u)$ in $L$
      \State Add fill edges that make $N_H(u)$ a clique
      \State Remove $u$ and its incident edges from $H$
    \EndWhile
    \State Create one tree node for every recorded bag $X_u$
    \State In reverse elimination order, connect each non-root bag $X_u$ to a
    later bag that contains $X_u\setminus\{u\}$
    \Return the constructed decomposition $\mathcal{T}$
  \end{algorithmic}
\end{algorithm}

The minimum modulator for the fixed decomposition is obtained from the binary program
\begin{align}
  \min_{z\in\{0,1\}^{V}}\quad & \sum_{v\in V}z_v \label{eq:tmgtd}\\
  \text{subject to}\quad
  & \sum_{v\in X_t}(1-z_v)\leq\eta+1
    && \text{for every }t\in V(T). \nonumber
\end{align}
Here $z_v=1$ means $v\in\mathcal{M}_{\eta}$. The objective therefore returns
the minimum modulator for the given decomposition, not an approximation to it.
We solve this MILP exactly. We use target width $\eta=10$ for MIS and MVC and
$\eta=6$ for MaxCut. The vertex-deletion version used here follows
\citet{Li2026Taming}; an edge-deletion variant on a fixed decomposition was
studied by \citet{Marchand2021TreeDR}.

For completeness, deleting $\mathcal{M}_{\eta}$ from every bag preserves the
tree-decomposition axioms for $G\setminus\mathcal{M}_{\eta}$. Vertex coverage
holds for every remaining vertex, every remaining edge stays covered by its
original bag, and the bags containing any remaining vertex are unchanged and
thus still connected. The MILP constraints then bound every remaining bag by
$\eta+1$ vertices.

\begin{figure}[!ht]
  \centering
  \hspace{-1.5em}
  \includegraphics[width=0.508\linewidth]{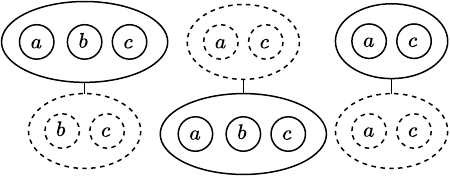}
  \caption{Three valid bag-pruning configurations. A dashed bag is removed
  when an adjacent bag contains it; reconnecting its neighbors through the
  containing bag preserves the tree-decomposition axioms.}
  \label{fig:bag-pruning}
\end{figure}

We also prune a bag when it is contained in an adjacent bag and reconnect its
neighbors through the containing bag. This preserves vertex and edge coverage;
for every vertex appearing on both sides of the removed bag, the containing
bag preserves the running-intersection property. Figure~\ref{fig:bag-pruning}
shows the three local configurations used by this operation.

\subsection{Advice Validity and Boundary Propagation}
\label{sec:advice-validity}
The advice used in our experiments is the restriction of a complete valid
neural solution to $\mathcal{M}_{\eta}$, so it always admits that neural
solution as a completion. The dynamic program in \textsc{N-FPT} also propagates the constraints created at the
boundary of the modulator. For MIS, every selected modulator vertex forces its
neighbors outside the modulator to be excluded. For MVC, every excluded
modulator vertex forces its neighbors outside the modulator to be selected.
MaxCut permits every binary assignment on the modulator and needs no analogous
repair. These rules make the DP enumerate exactly
$\mathcal{F}(G\mid a)$. If advice is supplied by another source and is invalid
within the modulator, a task-specific repair is applied before exact completion; the
guarantees then apply to the repaired advice.

\subsection{Consolidated-Prediction Variants}
\paragraph{\textsc{NIC-FPT}.}
\label{sec:icl-details}
At each round, the neural sampler draws $K$ valid completions
$S^{(1)},\ldots,S^{(K)}$ conditioned on the current consensus. For an
undecided vertex $v$, its selection count is
\[
  c_v=\sum_{j=1}^{K}\mathbf{1}\{v\in S^{(j)}\}.
\]
\textsc{NIC-FPT} fixes $v$ as selected when $c_v>\kappa$ and excludes
its neighbors. We use five rollout rounds with $K=20$ and
$\kappa=0.8K$; the first four rounds update the consensus and the fifth
completes the solution. Because
$\kappa>K/2$ and every sampled MIS solution is an independent set, two adjacent
vertices cannot both pass the threshold: their counts sum to at most $K$.
The consensus is therefore valid and conditions the next sampling round.
The final valid neural sample supplies modulator advice $a$;
\textsc{NIC-FPT} returns its best achievable completion $V_G(a)$.

\paragraph{\textsc{NRD-FPT}.}
Starting from a complete valid neural sample, \textsc{NRD-FPT} independently
defers each selected vertex with probability $\rho=0.8$. Retained selections
and their implied neighbor exclusions stay fixed; the sampler resamples a
valid completion over the other vertices. We use five rollout rounds,
applying deferral after the first four. The final valid neural sample
supplies modulator advice $a$, and \textsc{NRD-FPT} returns its best
achievable completion $V_G(a)$.

\section{Practical Runtime Performance}
We report empirical runtime results for the neural solver and exact completion
in \textsc{N-FPT}. For the neural solver, we reuse the GFlowNet implementation
from \citet{zhang2023letflowstellsolving}. The cost of exact completion depends
on the number of bag assignments considered by the dynamic program.

Following the prior metrics, we compute runtime per dataset, with each dataset processed in mini-batch. These batches serve as inputs to the training models during evaluation. We present the time consumption of each component separately to show their individual contributions.

\begin{table}[!ht]
  \centering
  \setlength{\tabcolsep}{0.0ex}
  \renewcommand{\arraystretch}{0.9}
  \caption{Runtime (seconds per dataset) of the neural and exact-completion
  components of \textsc{N-FPT}. One-time preprocessing is excluded.}
  \label{tab:runtime}

  {\scshape\footnotesize
    \begin{adjustbox}{max width=\columnwidth}
      \begin{tabular}{@{}l*{4}{@{\hspace{2ex}}c@{\hspace{1ex}}c}@{}}
        \toprule
        \multirow{2}{*}{Dataset}
          & \multicolumn{2}{c}{SS} & \multicolumn{2}{c}{SD}
          & \multicolumn{2}{c}{LS} & \multicolumn{2}{c}{LD} \\
        \cmidrule(lr){2-3}\cmidrule(lr){4-5}\cmidrule(lr){6-7}\cmidrule(lr){8-9}
          & Neural & Exact & Neural & Exact
          & Neural & Exact & Neural & Exact \\
        \midrule
        \textsc{Reg} & 95.79 & 147.48 & 60.33 & 39.30 & 159.04 & 183.29 & 102.76 & 53.54 \\
        \textsc{ER} & 44.40 & 43.09 & 26.14 & 18.38 & 71.37 & 39.52 & 44.38 & 21.36 \\
        \textsc{HK} & 113.52 & 498.87 & 78.41 & 248.59 & 234.93 & 776.04 & 163.07 & 343.75 \\
        \textsc{BA} & 128.30 & 712.77 & 63.82 & 134.72 & 266.63 & 1335.09 & 131.62 & 218.33 \\
        \textsc{WS} & 49.70 & 265.10 & 38.65 & 165.46 & 85.82 & 366.77 & 57.87 & 243.94 \\
        \bottomrule
      \end{tabular}
    \end{adjustbox}
  }
\end{table}

Another relevant consideration is the time required to construct the tree decomposition and compute its minimum modulator. In our experiments, this preprocessing takes approximately {\em one second} per instance on average. We exclude it from the component-level runtimes above because it is a one-time initialization cost and is negligible relative to the reported total.

\section{Related Work} \label{sec:related_work}

\paragraph{Neural Solvers for Graph Combinatorial Optimization.}
Neural solvers span two inference patterns. {\it One-shot} solvers, including
GNN-based neural solvers \citep{Schuetz_2022, gasse2019exact, wang2023learningbranchcombinatorialoptimization}, attention-based models \citep{vinyals2015pointer, kool2018attention} and diffusion models \citep{sun2023difusco, sanokowski2024diffusion, yu2024disco}, directly output solution in a single or several steps. Some further improve the results via MCTS approaches \citep{han2023gnn, fu2021generalize}. These are trained supervised or via handcrafted objectives.
They often learn latent representations that enable graph combinatorial optimization problems to be formulated as a vertex classification task. Despite strong inference-time performance, solution quality often degrades, especially for GNN-only models \citep{Angelini2022ModernGN, Gamarnik2023BarriersFT}.

A second approach frames graph combinatorial optimization as a Markov decision process (MDP), making RL a natural option \citep{khalil2017learning,bello2016neural, Ahn2020LearningWT}. RL-based neural solvers sequentially complete solutions through vertex decisions. These models use different neural architectures as {\it state encoders}, such as Transformer \citep{zhu2019causal}, GNNs~\citep{li2018combinatorial}, RNNs~\citep{vinyals2015pointer}, etc. GFlowNet~\citep{bengio2023gflownetfoundations, bengio2021flow} enhances sampling through flow-based losses~\citep{pan2023better}, balancing solution quality and diversity.

RL-based approaches make vertex decisions autoregressively; some models generate multiple decisions simultaneously, incorporating modified environments and reward shaping to guide learning. However, RL-based methods can suffer policy collapse~\citep{moalla2024no} and credit assignment difficulties~\citep{sutton1984temporal}, limiting exploration. Despite their successes, neural solvers can struggle with solution-quality guarantees and generalization. \textsc{N-FPT} uses their decisions as advice and certifies the best achievable completion under that advice.

\paragraph{Parameterized Algorithms.}
The complexity of graph problems is often measured by the number of vertices and edges. Parameterized algorithms also account for a chosen parameter~\citep{fomin2019kernelization}, such as optimal solution size, treewidth~\citep{bertele1973non, halin1976s,Robertson1984GraphMI}, or cutwidth~\citep{thilikos2005cutwidth}. FPT algorithms have polynomial dependence on input size for a fixed parameter value. Structural parameters such as treewidth can also reveal which parts of an instance contribute most to the computational difficulty.

\paragraph{Learning-Augmented Algorithms.}
Learning-augmented algorithms use predictions as advice to improve algorithmic performance. Much of the literature focuses on online algorithms~\citep{purohit2018improving}, though recent work extends to graph combinatorial optimization, including MaxCut~\citep{dong2025learningaugmentedstreamingalgorithmsapproximating} and independent set~\citep{braverman_et_al:LIPIcs.APPROX/RANDOM.2024.24}. This connection with predictive models has also motivated practical applications~\citep{shin2023improved, dinitz2022algorithms}. The advice may be imperfect; algorithmic guarantees specify how the result depends on its quality.

\section{Future Directions}
Additional neural models, such as {\it diffusion models}, can supply advice to \textsc{N-FPT}. Similarly, other parameterized algorithms can take the place of treewidth dynamic programming. For instance, if we replace treewidth dynamic programming with parameterized approximation schemes or randomized parameterized algorithms, we may obtain other learning-augmented algorithms with different runtime and solution-quality guarantees. The learning-augmented framework also motivates {\it beyond-worst-case} analysis combining structural assumptions, such as small treewidth modulators, with assumptions on neural advice.

\end{document}